\documentclass[10pt, a4paper]{article}
\usepackage{booktabs}
\usepackage[hide]{todo}

\usepackage{amsmath,amssymb,amsfonts}
\newcommand{\citepl}{\citeplanguageresource}
\newcommand{\citetl}{\citetlanguageresource}
\newcommand{\version}{final}%
\usepackage[\version]{lrec2026} %

\usepackage{ifthen}
\usepackage{amssymb}%
\usepackage{pifont}%
\usepackage[dvipsnames]{xcolor}

\usepackage{inconsolata}

\newcommand{\cmark}{\color{OliveGreen}{\ding{51}}}
\newcommand{\xmark}{\color{BrickRed}{\ding{55}}}
\newcommand{\rTodo}[1]{}

\title{Assessing the Political Fairness of Multilingual LLMs:\\A Case Study based on a 21-way Multiparallel EuroParl Dataset}
\name{Paul Lerner, François Yvon} 

\address{Sorbonne Université, CNRS, ISIR \\
         75005, Paris, France \\
         lerner@isir.upmc.fr, yvon@isir.upmc.fr}

\abstract{
  The political biases of Large Language Models (LLMs) are usually assessed by simulating their answers to English surveys. %
  In this work, we propose an alternative framing of political biases, relying on principles of fairness in multilingual translation. We systematically compare the translation quality of speeches in the European Parliament (EP), observing systematic differences with majority parties from left and right being better translated than outsider parties.
  This study is made possible by a new, 21-way multiparallel version of EuroParl, the parliamentary proceedings of the EP, which includes
  the political affiliations of each speaker. The dataset consists of 1.5M sentences for a total of 40M words and 249M characters. It covers three years, 1000+ speakers, 7 countries, 12 EU parties, 25 EU committees, and hundreds of national parties.
 \\ \newline  }

\begin{document}

\maketitleabstract

\section{Introduction}

\begin{figure}[t]
    \centering
    \includegraphics[width=\linewidth, trim={2cm 12.5cm 8cm 2.5cm},clip]{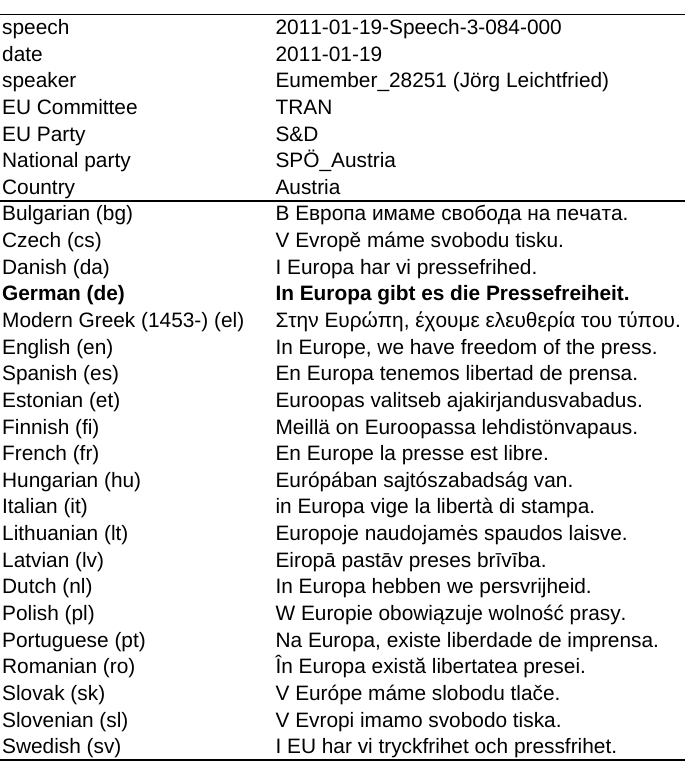}
    \caption{A fully worked-out example of 21-EuroParl: a German (\textsc{de}) source sentence is aligned to translations in 20 other languages. Metadata includes the speaker's political affiliation.}
    \label{fig:eg}
\end{figure}

\begin{figure}[t]
    \centering
    \includegraphics[width=\linewidth]{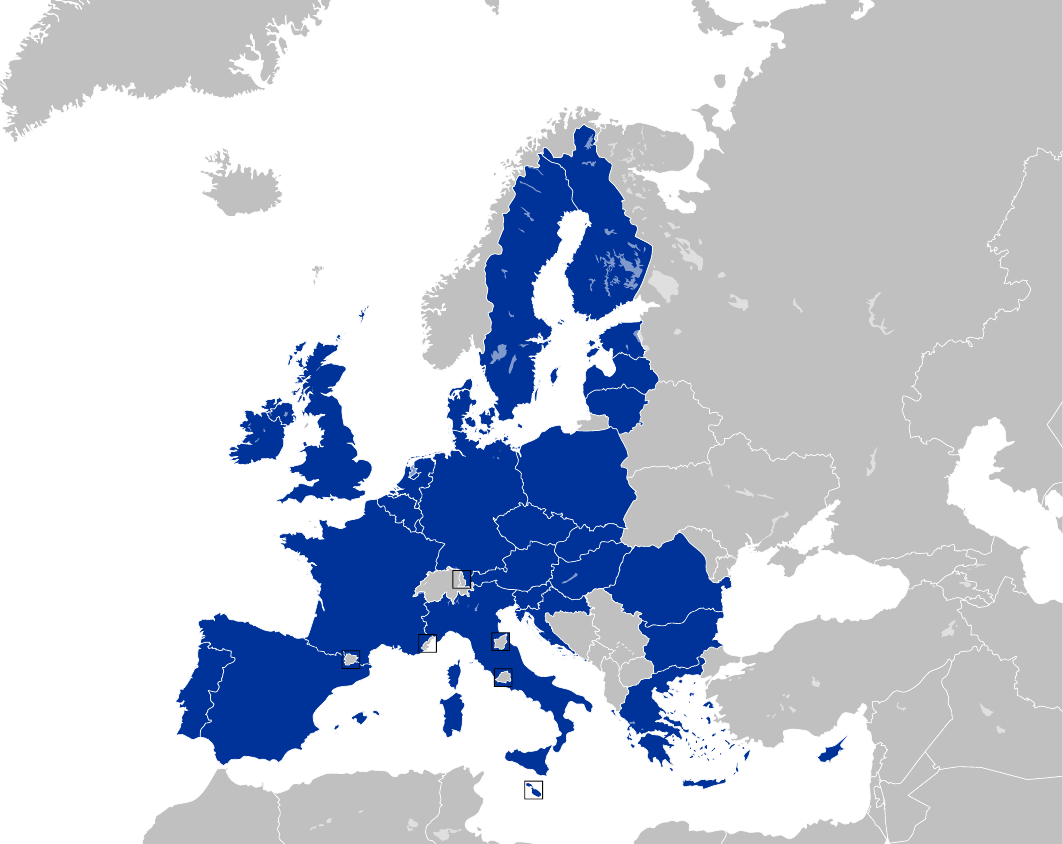}
    \caption{Countries covered in 21-EuroParl, members of the European Union in 2009-2011 (Kolja21, CC BY 3.0, via Wikimedia Commons)}
    \label{fig:2013_European_Union_map}
\end{figure}

Parliamentary debates have long been used as a major source of parallel texts in the history of Machine Translation \citepl{brown-etal-1990-statistical}. Notably, the EuroParl dataset \citepl{koehn-2005-europarl} enabled the study of an increasing diversity of language pairs as the European Union (EU) was growing and played a key part in WMT's shared tasks \citep{WMT:2006,bojar-etal-2015-findings,kocmi-EtAl:2022:WMT}.
Given their content, parliamentary debates are also relevant for various fields, including political sciences \citep{hoyland-etal-2014-predicting,deligiaouri2016analysis,schwalbach-etal-2024-mind,sebok2025Comparative}.
Unfortunately, existing corpora derived from the European Parliament's debates (including the noteworthy EuroParl\footnote{Also available via OPUS \citep{tiedemann-2012-parallel}.}) %
share two limitations (see Table~\ref{tab:vs_rw}):
\begin{itemize}
    \item they are biparallel, i.e., they pair texts in two languages, although speeches were originally translated into $n$ languages, which prevents any multilingual comparable analysis; %
    \item the associated metadata is incomplete. %
\end{itemize}
To fill this gap, we build on LinkedEP \citepl{noy-etal-2017-debates} to present 21-EuroParl, a 21-way multiparallel dataset aligned at the sentence level with extensive metadata, including information about the speaker's political affiliation (national and EU party),
see example on Figure~\ref{fig:eg}. The dataset contains a grand total of 72,234~instances and
spans over three years, from 2009 to 2011 (included). It comprises speeches from 1000+ speakers coming from 27 countries (see Figure~\ref{fig:2013_European_Union_map}), 12 EU parties, 25 EU committees, and hundreds of national parties.

\begin{table*}[t]
    \centering
    \resizebox{\textwidth}{!}{
    \begin{tabular}{lrcccccc}
    \toprule
        \textbf{Dataset} & \textbf{\# Lang. Pairs} & \textbf{Sent. Align.} & \textbf{Multiparallel}  & \textbf{EU Committee} & \textbf{EU Party} & \textbf{National Party} & \textbf{OL}\\
        \midrule
        LinkedEP & 420 & \xmark & \cmark & \cmark & \cmark & \cmark& \cmark \\
        MPDE & 42 & \xmark &  \cmark& \xmark & \xmark & \xmark& \cmark \\
         DCEP & 253 & \cmark & \xmark &\xmark  & \xmark &\xmark & \xmark \\
        EuroParl-UdS & 2 & \cmark & \xmark & \cmark & \cmark & \cmark& \cmark \\
        EuroParl v10  & 420 & \cmark & \xmark &\xmark  & \cmark &\xmark& \cmark \\
        Europarl-ST & 12 & \cmark& \cmark & \cmark & \cmark & \cmark & \cmark \\
        21-EuroParl & 420  & \cmark& \cmark & \cmark & \cmark & \cmark & \cmark\\
        \bottomrule
    \end{tabular}}
    \caption{21-EuroParl compared to related parallel datasets. OL: Original Language.}
    \label{tab:vs_rw}
\end{table*}

To showcase the potential of 21-EuroParl, we conduct a study on the \emph{political fairness} of multilingual Large Language Models (LLMs). 
This question is critical as LLMs are used daily through chatbots by hundreds of millions of users \cite{milmo2023chatgpt,similarweb} and are increasingly embedded in democratic procedures and processes \cite{DBLP:journals/corr/abs-2306-11932,tessler-etal-2024-ai,revel2025ai}. 
However, their biases have so far mostly been assessed through simulated surveys, written in English %
\citep{feng-etal-2023-pretraining,rozado-etal-2023-political, hartmann-etal-2023-political,santurkar-etal-2023-whose,durmus-etal-2024-measuring,motoki-etal-2024-more,potter-etal-2024-hidden,boelaert-etal-2024-how,rottger-etal-2024-political,ceron-etal-2024-prompt}. 
We propose an alternative framing of political biases and ask the following question: \emph{does LLM translation quality systematically vary depending on the political content of the source text?} 
The answer is not straightforward as existing Machine Translation metrics are not well calibrated across language pairs.
We therefore introduce a novel way to aggregate translation scores across multiple language pairs and partisan
speeches based on Borda counts.
We find systematic differences in scores depending on the source languages, as well as
the party affiliation, across several model sizes and families.
In summary, our contributions are threefold:
\begin{itemize}
    \item a new, 21-way multiparallel version of
EuroParl, including the political affiliation of speakers (Section~\ref{sec:dataset});
\item a new  Borda-based method  to aggregate rankings across language pairs (Section~\ref{sec:Experiments});
\item observations regarding the politically unfair translation of LLMs (Section~\ref{sec:results}).
\end{itemize}

We make our dataset, code, and translation outputs available for reproducibility\footnote{\url{https://github.com/PaulLerner/21-EuroParl}}.

\section{Related Work \label{ssec:related}}
\subsection{From Parliamentary Speeches into Multilingual Corpora}

The preparation and release of EuroParl based on the processing of speeches at the EP \citepl{koehn-2005-europarl} has had a major impact on machine translation for European languages. Since then, multiple works have, like ours, extended EuroParl with additional information or metadata, such as the original source language or the political affiliation of speakers. For instance, CoStEP \citepl{graen2014cleaning} includes multiple automatic linguistic annotations enabling various types of cross-lingual corpus analyses. However, they are often limited to a small number of languages, see Table~\ref{tab:vs_rw}.
EuroParl-UdS \citepl{karakanta2018europarl} extends EuroParl with metadata about the original source language and the political affiliation of the speaker, focusing on only three languages.
MPDE \citepl{amponsah-kaakyire-etal-2021-not} augments EuroParl with information about the original source language and about the use of pivot translation through a third language (e.g. \textsc{fr-en} then \textsc{en-de} instead of \textsc{fr-de} directly). As this information is not explicitly available, they approximate it by considering all translations from 2004 onward as performed through a pivot language. Moreover, they are limited to 7 languages and do not align texts at the sentence level. \citetl{iranzo-sanchez-etal-2020-europarlst} focus on speech translation and, like us, rely on LinkedEP to collect metadata. However, their study is also limited to a subset of four languages, dubbed Europarl-ST.

The undocumented EuroParl v10 now includes speaker metadata, notably their EU party affiliation.\footnote{\url{https://www.statmt.org/europarl/v10/}} However, it is lacking the mention of the relevant EU committee as well as the national party affiliation. Moreover, it is only biparallel, which prevents comparable analysis across languages.

Related to parliamentary debates, the DCEP dataset covers various documents from the European Parliament, such as press releases \citepl{hajlaoui-etal-2014-dcep}. However, to avoid overlap with EuroParl, it does not contain the verbatim reports of speeches made in  the European Parliament's plenary. Moreover, it is only biparallel and does not include metadata.

Finally, \citetl{yang-etal-2023-multieup} propose Multi-EuP to assess the multilingual fairness of Information Retrieval methods, for which they consider topics of the EP as queries and relevant documents as any speech from this topic.

\subsection{Political Biases of LLMs}
\paragraph{English surveys} To the best of our knowledge, this work is the first to leverage Machine Translation as a way to question the political biases of multilingual LLMs. Indeed, most existing works on political biases: (a) focus on monolingual, English-centric LLMs; %
(b) rely on surveys ``administered'' to LLMs (e.g., querying a model about a degree of agreement with statements such as ``\textit{Sex outside marriage is usually immoral}'').
More precisely, we refer to the following works: \citep{feng-etal-2023-pretraining,rozado-etal-2023-political,santurkar-etal-2023-whose,motoki-etal-2024-more,potter-etal-2024-hidden}. 

\paragraph{Brittleness} Following works have demonstrated the brittleness of the survey approach, as the computation of LLMs' answers is sensitive to the exact phrasing of the prompt, also yielding a lack of overall coherence \citep{rottger-etal-2024-political,ceron-etal-2024-prompt,boelaert-etal-2024-how}.

\paragraph{Multilingual} \citet{durmus-etal-2024-measuring,helwe-etal-2025-navigating} rely on MT to translate English surveys in multiple languages. However, this poses a confound since MT may (i) obviously contain errors, including stance-changing errors \citep{nazanin}; (ii) be biased towards a political party, as we will find out.
Nevertheless, \citet{durmus-etal-2024-measuring} find that surveying LLMs in English, Russian, Chinese, or Turkish always leads to occidental responses, i.e. the answers are consistent across languages. 
On the contrary, \citet{helwe-etal-2025-navigating} finds that ``language has a strong influence'', although they do not control for the spurious prompt effects described above.
\citet{leo} have jointly assessed the variation caused by language and prompt phrasing. They found that  the effect of language was question-dependent, i.e., for a subset of questions the answer is consistent across languages while for another subset the answer differs across languages.

\paragraph{Other Methods} 
In addition to their survey-based method, \citet{potter-etal-2024-hidden} conduct a human-computer interaction experiment. They find that after debating with an LLM in English, Trump supporters increased their Biden-leaning.
\citet{ceron-etal-2025-what} study the political content of LLMs English pretraining and posttraining data. Using a left-right content classifier, they find a majority of left-leaning data, which concurs with the results discussed above.
\citet{rottger-etal-2025-issuebench} assess the stance of LLMs when prompted to generate English texts about debated issues. They find that LLMs align closely with US Democrats, as opposed to Republicans.

\paragraph{Fairness}
We depart from these works and, instead of measuring political biases of LLMs based on the detected political leaning of the text they produce, we frame it in terms of a \emph{group-fairness analysis}, based on a practical application
 (Machine Translation), and assess biases over a large set of 420 language pairs. This allows us to also cover a much more diverse set of texts, reflecting the fact that the political spectrum of the European Union is arguably more varied than that of the bipartisan US, which largely maps to a liberal-conservative axis \citep{ramaciotti-etal-2024-american,vendeville:hal-05222448}. In the EU, other ideological dimensions emerge, e.g., Green-Alternative-Libertarian \textsl{versus} Traditional-Authoritarian-Nationalist (aka GAL-TAN), Urban-Rural, or Pro-Anti Immigration \citep{jolly-etal-2022-chapel}.

 Our methodology thus borrows from the \textit{fairness} literature \citep{verma-rubin-2018-fairness,czarnowska-etal-2021-quantifying,liang-etal-2022-holistic,barocas-etal-2023-fairness,gallegos-etal-2024-bias}, as we treat political affiliation as a \emph{protected attribute} which models should not be sensitive to (instead of more commonly-used attributes such as gender \cite{savoldi-etal-2021-gender}). %

\begin{figure}[t]
    \centering
    \includegraphics[width=\linewidth]{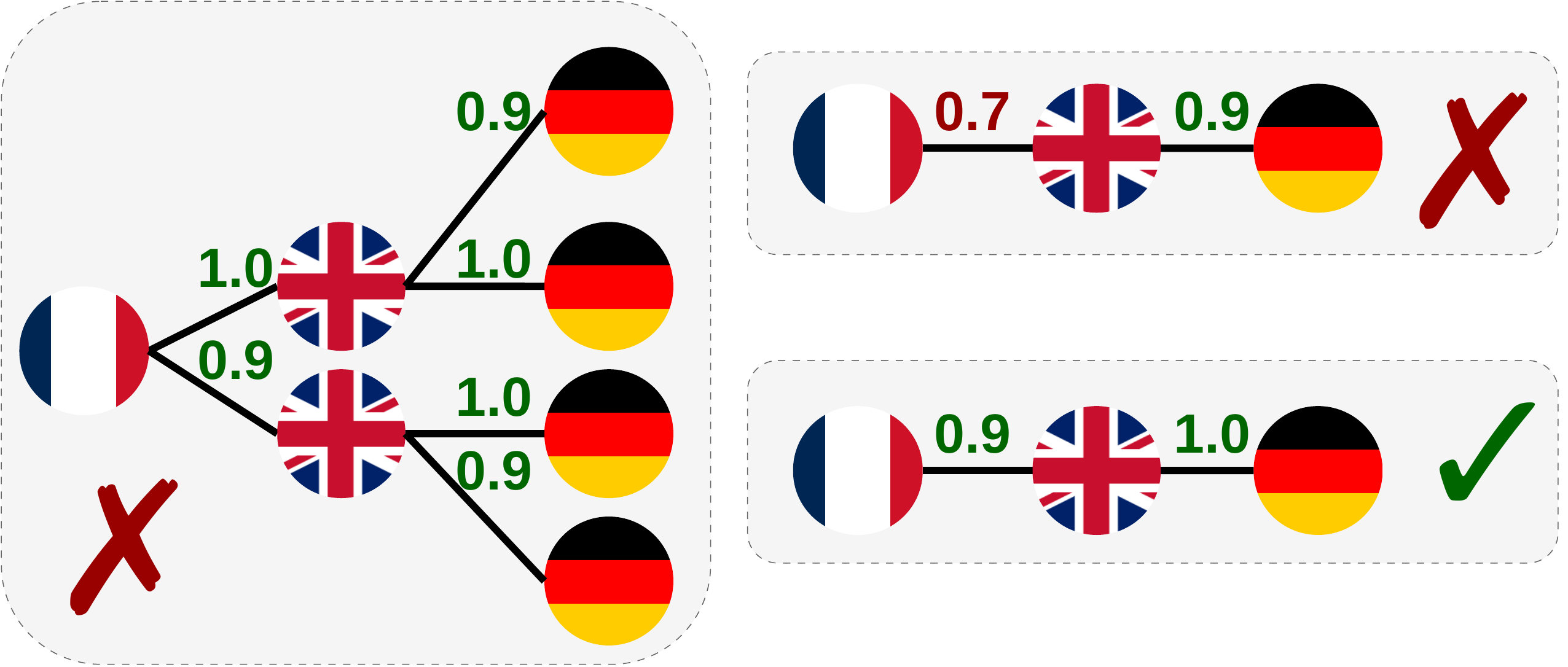}
    \caption{Illustration of various alignment scenarios: (left) a 1-\textsc{fr}-2-\textsc{en} alignment may result in a 1-\textsc{fr}-4-\textsc{de} alignment if each \textsc{en} sentence is itself aligned to 2 \textsc{de} sentences; (top, right) although \textsc{en-de} is well aligned, \textsc{fr-en} is not, so we reject the multialignment; (bottom, right) valid multialignment: all bialignments are 1-1 and have a score above the threshold of 0.8}
    \label{fig:alignment}
\end{figure}

\section{Preparing 21-EuroParl}\label{sec:dataset}%
\subsection{LinkedEP}

21-EuroParl heavily builds upon LinkedEP \citepl{noy-etal-2017-debates}, which represents the European Parliament's debates as a knowledge graph in which speeches, speakers, and political parties are nodes linked by relations.

As LinkedEP is stored in RDF triples, we query it using SPARQL to retrieve, for each speech, all translations as well as their metadata. 
LinkedEP contains 191,921 speeches, partially translated into 23 languages. However, \textsc{hr} (Croatian) and \textsc{mt} (Maltese) are hardly covered\footnote{%
  Croatian only acquired the status of an official language of the EU in 2013.
  Maltese and Irish were, until recently, subject to a derogatory status, likely due to a lack of trained interpreters/translators,
  and they were only partially supported.
  Ireland and Malta also have English as a co-official language.} so we exclude them to arrive at the 21 languages of 21-EuroParl: 
\done%
\textsc{bg}, 
\textsc{cs}, 
\textsc{da}, 
\textsc{de}, 
\textsc{el},   
\textsc{en}, 
\textsc{es}, 
\textsc{et}, 
\textsc{fi}, 
\textsc{fr}, 
\textsc{hu}, 
\textsc{it}, 
\textsc{lt},  
\textsc{lv}, 
\textsc{nl}, 
\textsc{pl}, 
\textsc{pt},  
\textsc{ro}, 
\textsc{sk}, 
\textsc{sl}, and  
\textsc{sv} (ISO 639-1 codes). %

Not all the 191,921 speeches are available in 21 languages: we found that many texts tagged as translations in LinkedEP were in fact duplicates of the source speech. After cleaning using fastText's language identification tool \citep{joulin-etal-2017-bag}, we were left with 37,799 speeches that were available in 21~languages. Less than a thousand were then filtered out because of missing metadata. %
\done%
The next important step was to compute a sentence alignment.

\subsection{Sentence Alignment \label{ssec:sentence-alignment}}

Sentence alignment was performed with Bertalign \citep{liu-etal-2023-bertalign}, which uses LaBSE sentence embeddings \citep{feng-etal-2022-languageagnostic}. After initial experiments, we replaced Bertalign's original sentence segmenter\footnote{\url{https://github.com/mediacloud/sentence-splitter}} by Trankit's\footnote{\url{github.com/nlp-uoregon/trankit}} %
\citep{nguyen2021trankit}.

Bertalign computes alignments for pairs of languages (bialignments). We first performed English-centric bialignments, from English to the 20 other languages, yielding 20 bialignments per speech.
To control the alignment quality and avoid one-to-many alignments propagating into one-to-many-many (see Figure~\ref{fig:alignment}), we only kept examples with one-to-one alignments in the 20 bialignments. Furthermore, we filtered out examples if any of their bialignment had a Bertalign score (LaBSE's cosine similarity between the source and target sentences) of less than $0.8$. 

The entire process yielded 72,234 examples (1,516,914 sentences) aligned in 21~languages, a reduction of about 76\% from the original LinkedEP. 

We will make alignments available so that the dataset can easily be extended with less conservative thresholds or one-to-many alignments.

\begin{table}[t]
    \centering
    \begin{tabular}{lr}
    \toprule
    \textbf{Metadata} & \textbf{Unique Values} \\
        \midrule
      Speech & 27,873 \\
Speaker & 1,094 \\
National party & 272 \\
EU party & 12 \\
EU committee & 25 \\
Country & 27 \\ %
        \bottomrule
    \end{tabular}
    \caption{Summary of the available metadata in 21-EuroParl}
    \label{tab:metadata}
\end{table}

\subsection{Overview of the dataset}
\begin{table}[t]
\centering
\begin{tabular}{lrr}
\toprule
\textbf{Language} & \textbf{\# Words} & \textbf{\# Chars} \\
 \midrule
\textsc{bg} & 2,057,994 & 12,032,635 \\
\textsc{cs} & 1,778,026 & 10,668,881 \\
\textsc{da} & 1,949,508 & 11,414,059 \\
\textsc{de} & 2,027,728 & 13,216,642 \\
\textsc{el} & 2,140,104 & 13,126,027 \\
\textsc{en} & 2,088,541 & 11,541,038 \\
\textsc{es} & 2,240,743 & 12,706,854 \\
\textsc{et} & 1,524,824 & 10,738,616 \\
\textsc{fi} & 1,487,870 & 11,628,885 \\
\textsc{fr} & 2,213,235 & 13,027,959 \\
\textsc{hu} & 1,802,043 & 12,171,311 \\
\textsc{it} & 2,027,404 & 12,546,078 \\
\textsc{lt} & 1,651,547 & 10,893,475 \\
\textsc{lv} & 1,682,997 & 10,807,953 \\
\textsc{nl} & 2,119,408 & 12,763,061 \\
\textsc{pl} & 1,781,796 & 11,921,243 \\
\textsc{pt} & 2,182,858 & 12,322,265 \\
\textsc{ro} & 2,102,942 & 12,336,509 \\
\textsc{sk} & 1,777,101 & 10,948,052 \\
\textsc{sl} & 1,792,083 & 10,570,973 \\
\textsc{sv} & 1,847,453 & 11,381,764 \\
 \midrule
\textbf{Total} & \textbf{40,276,205} & \textbf{248,764,280} \\
\bottomrule
\end{tabular}
\caption{Total word and character counts in the 72,234 sentences for each 21-EuroParl language}
\label{tab:word_stats}
\end{table}

21-EuroParl spans over three years, %
from 2009 to 2011 (included). While we aligned the speeches at the sentence level, we kept the speech identifier of each example along with the date of the debate and the language in which the speech was originally uttered. Importantly, we also have a unique identifier for the 1000+ speakers in the dataset, along with their political affiliation: national party (and corresponding country), EU party, and EU committee (see Figure~\ref{fig:eg}). The dataset covers 27 countries (see Figure~\ref{fig:2013_European_Union_map}), 12 EU parties, 25 EU committees, and hundreds of national parties. These numbers are summarized in Table~\ref{tab:metadata}. The entire dataset contains 40,276,205 words\footnote{Word tokenization is performed using spacy’s multilingual tokenizer \url{https://spacy.io/models/xx}.}\done%
and 248,764,280 characters, a breakdown per language is in Table~\ref{tab:word_stats}.

\begin{figure}[t]
    \centering
    \includegraphics[width=\linewidth]{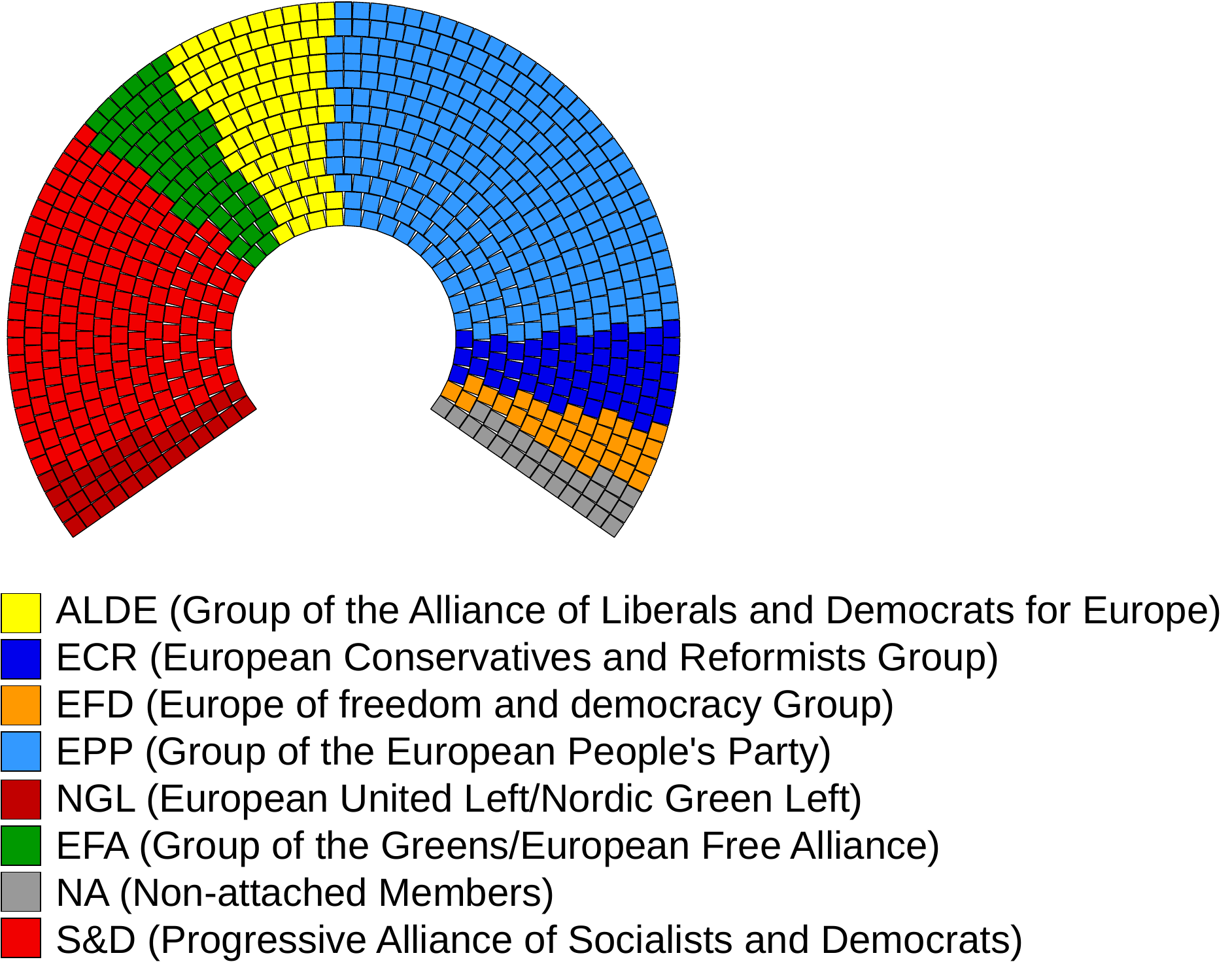}
    \caption{Composition of the 7$^{\text{th}}$ European Parliament, representing EU parties in 21-EuroParl, (Ssolbergj, CC BY-SA 3.0, via Wikimedia Commons)}
    \label{fig:eu_parliament}
\end{figure}

We split the dataset chronologically, reserving year 2009 for training, 2010 for validation, and 2011 for testing. All results in this study are computed on the test set of 23,386 examples; the split is provided to facilitate future research. The eight EU parties represented in the test set are displayed in Figure~\ref{fig:eu_parliament}. %

\begin{table*}[t]

\resizebox{\textwidth}{!}{\begin{tabular}{lrrrrrrrrrrrrrrrrrrrrr}
\toprule
src/tgt & \textsc{bg} & \textsc{cs} & \textsc{da} & \textsc{de} & \textsc{el} & \textsc{en} & \textsc{es} & \textsc{et} & \textsc{fi} & \textsc{fr} & \textsc{hu} & \textsc{it} & \textsc{lt} & \textsc{lv} & \textsc{nl} & \textsc{pl} & \textsc{pt} & \textsc{ro} & \textsc{sk} & \textsc{sl} & \textsc{sv} \\
\midrule
\textsc{bg} & -- & 26.3 & 28.6 & 27.9 & 30.4 & 44.6 & 38.6 & 15.6 & 17.7 & 32.3 & 20.9 & 28.1 & 18.0 & 19.5 & 26.8 & 24.3 & 32.8 & 33.1 & 24.6 & 25.5 & 28.4 \\
\textsc{cs} & 27.8 & -- & 27.2 & 27.0 & 27.6 & 41.2 & 36.3 & 15.4 & 17.4 & 30.8 & 20.5 & 26.7 & 16.7 & 18.1 & 26.4 & 23.9 & 30.7 & 30.5 & 31.2 & 8.2 & 27.3 \\
\textsc{da} & 27.5 & 24.3 & -- & 27.7 & 27.2 & 44.2 & 36.9 & 15.0 & 17.9 & 31.6 & 19.7 & 27.2 & 16.0 & 17.8 & 26.9 & 22.2 & 31.0 & 30.7 & 22.6 & 23.3 & 32.6 \\
\textsc{de} & 25.1 & 25.0 & 28.7 & -- & 26.6 & 43.2 & 37.2 & 15.3 & 17.7 & 31.5 & 20.0 & 27.1 & 16.3 & 17.5 & 28.6 & 22.4 & 30.9 & 31.1 & 23.4 & 23.9 & 28.5 \\
\textsc{el} & 20.2 & 14.7 & 22.5 & 23.4 & -- & 45.3 & 37.7 & 10.5 & 13.9 & 31.3 & 12.7 & 26.5 & 10.9 & 11.8 & 19.3 & 15.1 & 29.4 & 30.1 & 16.2 & 21.2 & 22.3 \\
\textsc{en} & 37.0 & 32.2 & 37.5 & 36.2 & 38.8 & -- & 50.1 & 19.0 & 22.2 & 40.4 & 25.2 & 34.6 & 20.4 & 23.9 & 32.3 & 28.7 & 41.7 & 44.3 & 30.3 & 31.2 & 37.5 \\
\textsc{es} & 27.8 & 24.6 & 29.3 & 21.8 & 30.8 & 48.7 & -- & 15.7 & 18.1 & 34.1 & 20.0 & 30.0 & 17.1 & 18.2 & 27.2 & 23.7 & 36.1 & 35.4 & 24.2 & 24.6 & 29.3 \\
\textsc{et} & 23.9 & 21.7 & 23.9 & 23.7 & 23.6 & 34.9 & 32.5 & -- & 17.4 & 27.3 & 18.9 & 23.6 & 15.5 & 16.7 & 23.4 & 20.3 & 26.9 & 26.4 & 20.0 & 19.9 & 24.1 \\
\textsc{fi} & 23.5 & 21.4 & 24.3 & 23.7 & 24.0 & 35.6 & 32.4 & 15.1 & -- & 27.5 & 18.8 & 23.8 & 14.8 & 16.0 & 24.0 & 19.8 & 27.2 & 26.4 & 19.8 & 20.0 & 25.5 \\
\textsc{fr} & 25.0 & 23.4 & 27.1 & 26.7 & 28.3 & 42.8 & 38.8 & 13.0 & 16.2 & -- & 19.6 & 29.0 & 14.9 & 16.0 & 26.8 & 22.2 & 33.0 & 32.0 & 21.8 & 22.8 & 27.2 \\
\textsc{hu} & 24.1 & 21.7 & 24.3 & 23.8 & 24.1 & 36.1 & 33.2 & 13.6 & 16.0 & 28.1 & -- & 24.4 & 15.0 & 16.1 & 24.1 & 20.3 & 28.4 & 26.6 & 19.9 & 20.5 & 24.5 \\
\textsc{it} & 22.7 & 21.9 & 25.4 & 22.0 & 25.6 & 39.9 & 36.2 & 13.6 & 13.5 & 31.5 & 18.1 & -- & 14.3 & 15.9 & 25.8 & 21.1 & 31.1 & 29.9 & 21.0 & 20.9 & 25.6 \\
\textsc{lt} & 23.0 & 20.1 & 22.2 & 21.7 & 21.8 & 33.2 & 30.7 & 13.6 & 14.6 & 26.0 & 16.9 & 22.2 & -- & 17.1 & 21.7 & 19.1 & 25.5 & 24.9 & 18.8 & 19.3 & 22.6 \\
\textsc{lv} & 24.0 & 21.7 & 24.1 & 23.4 & 23.6 & 35.9 & 32.7 & 14.9 & 15.8 & 27.2 & 18.5 & 23.4 & 17.0 & -- & 23.1 & 20.5 & 27.2 & 26.6 & 20.0 & 20.5 & 24.0 \\
\textsc{nl} & 24.3 & 21.9 & 25.4 & 25.6 & 24.9 & 37.4 & 33.8 & 13.5 & 16.1 & 29.2 & 18.6 & 25.5 & 15.0 & 16.0 & -- & 20.5 & 28.7 & 27.3 & 20.2 & 20.9 & 25.8 \\
\textsc{pl} & 26.2 & 24.0 & 25.3 & 21.7 & 26.1 & 38.1 & 34.6 & 14.4 & 16.2 & 29.4 & 19.3 & 25.5 & 16.2 & 17.3 & 24.7 & -- & 29.4 & 28.6 & 22.5 & 22.7 & 25.7 \\
\textsc{pt} & 29.0 & 25.2 & 28.5 & 26.1 & 30.8 & 46.8 & 41.6 & 15.2 & 17.6 & 34.3 & 20.4 & 29.9 & 16.8 & 18.1 & 27.3 & 23.3 & -- & 33.9 & 23.4 & 23.9 & 28.4 \\
\textsc{ro} & 29.6 & 25.8 & 29.6 & 25.5 & 32.3 & 48.7 & 42.1 & 15.5 & 17.5 & 34.8 & 21.2 & 30.0 & 17.1 & 19.1 & 27.7 & 23.5 & 35.7 & -- & 24.3 & 24.7 & 29.7 \\
\textsc{sk} & 28.4 & 32.1 & 27.6 & 27.8 & 27.8 & 42.0 & 37.0 & 14.8 & 17.1 & 31.2 & 19.9 & 26.9 & 16.6 & 18.5 & 26.4 & 24.3 & 31.1 & 31.0 & -- & 24.9 & 27.4 \\
\textsc{sl} & 27.7 & 25.7 & 27.1 & 27.2 & 26.9 & 41.2 & 36.6 & 14.5 & 16.6 & 30.7 & 19.4 & 26.4 & 16.8 & 18.0 & 25.9 & 23.3 & 30.4 & 30.4 & 24.7 & -- & 27.0 \\
\textsc{sv} & 26.3 & 23.4 & 31.4 & 26.5 & 26.3 & 42.1 & 35.7 & 14.9 & 17.4 & 30.7 & 19.1 & 26.6 & 15.7 & 17.2 & 26.3 & 21.7 & 30.1 & 29.7 & 21.5 & 22.1 & -- \\
\bottomrule
\end{tabular}
}
    \caption{sBLEU scores for Llama-3.1-70B: source languages in rows and target languages in columns. Scores are comparable within the same column (target language) but not across columns (within the same row, i.e. source language).}
    \label{tab:sBLEU scores of Llama-3.1-70B}
\end{table*}

\section{Experiments}\label{sec:Experiments}
\subsection{Method}\label{ssec:Method}
\paragraph{Translation fairness} As a first study, we would like to assess whether the quality of translations provided by multilingual LLMs varies depending on the EU party, \footnote{We focus only on EU parties, given the exceedingly large number of national parties.} which may indicate a political bias of the model.  We thus generate automatic translations of each speech, considering all possible directions of the 21-way multiparallel dataset, that is $21 \times 20=420$ language pairs.\done%

\paragraph{Metrics}
Translation quality is estimated with two standard reference-based metrics: sBLEU\footnote{\texttt{nrefs:1|case:mixed|eff:yes|tok:13a|} \texttt{smooth:exp|version:2.5.1} \citep{post-etal-2018-call}}, i.e. per-sentence BLEU \citep{papineni-etal-2002-bleu,chen-cherry-2014-systematic}, %
and COMET\footnote{\texttt{wmt22-comet-da}} \citep{rei-etal-2022-comet22}.  %
It is however well-known that these metrics are not well calibrated across language pairs \citep{bugliarello-etal-2020-easier,zouhar-etal-2024-pitfalls} and that the scores cannot be easily compared.
More precisely, COMET scores are not well calibrated across language pairs \citep{zouhar-etal-2024-pitfalls}, likely because it inputs not only the hypothesis and reference, but also the source sentence. sBLEU, while also not well calibrated across target languages due e.g.\ to tokenizations issues, still enables to compare multiple source languages translating into the same target language (Section~\ref{ssec:Language Fairness}), as it only relies on a reference and a hypothesis in that target language.

\paragraph{Borda Count} Therefore, to measure political fairness based on sBLEU and COMET, we compute a Borda Count of EU parties per language pair (Section~\ref{ssec:Political Biases}). Borda Count is a simple, yet effective, way to aggregate rankings \citep{mclean2019a}. For a given model, metric, and language pair, all EU parties are ranked according to the metric score. The EU party with the lowest score gets 0 points, the second-to-last gets 1 point, etc. until the best one gets 7 points (as there are 8 parties).
While initially proposed as a voting system, Borda Counts has been used in Information Retrieval to aggregate rankings \cite{DBLP:conf/sigir/AslamM01,ranx.fuse}. It has also been used more recently for benchmarking models across tasks with different metrics \citep{colombo-etal-2022-what} or meta-evaluating metrics across language pairs \citep{freitag-etal-2023-results}.
Our method differs from the usual fairness metrics, e.g. demographic parity or group calibration, that are designed for classification settings \citep{czarnowska-etal-2021-quantifying,barocas-etal-2023-fairness}.

\subsection{Implementation}
We experiment with three families of posttrained LLMs: Gemma-3-4B \citep{team-etal-2025-gemma}, Qwen3 \citep{yang-etal-2025-qwen3}, in two sizes (4B and 8B), and Llama-3.1 \citep{llamateam-etal-2024-llama}, in two sizes as well (8B and 70B). 

We originally carried out experiments with the smaller Llama-3.2 (in sizes 1B  and 3B) but found the translation scores overall too poor to be reliably compared across parties (for most language pairs, sBLEU scores were below 10, see Appendix~\ref{a:sbleu}). Llama-3.1 is a multilingual model trained with a very large share of English texts, accounting for 92\% of the pretraining data (or 75\% when excluding code) and 97\% English posttraining data (or 82\% when excluding code). Little is known about Gemma or Qwen3's data mix, except that Qwen3's covers 119 languages. %

We use a standard prompt template for zero-shot translation: \texttt{"<src\_lang>: <src\_text>\textbackslash{}n<tgt\_lang>:"} where \texttt{<src\_lang>} and \texttt{<tgt\_lang>} are replaced by the English name of the source and target languages, respectively, and \texttt{<src\_text>} is replaced by the source sentence. For example: \texttt{"English: In Europe, we have freedom of the press.\textbackslash{}nFrench:"}.\footnote{For Gemma, experiments with that prompt resulted in chatbot boilerplate answers, e.g.: \texttt{"Here are a few options for the English translation, with slightly different nuances:\textbackslash{}n\textbackslash{}n*   **Most straightforward:** I believe that [...]"}. We therefore changed the prompt to explicitly ask for the translation output only: \texttt{"You are a professional <src\_lang> to <tgt\_lang> translator. Please translate the following <src\_lang> text into <tgt\_lang>:\textbackslash{}n<src\_text>\textbackslash{}n\textbackslash{}nProduce only the <tgt\_lang> translation, without any additional explanations or commentary."}}
This prompt is then formatted as a user query using model-specific templates. %
We use the recommended decoding parameters, i.e. nucleus sampling with $p=0.9$ and temperature $\tau = 0.6$ \citep{Holtzman2020The}.
Models are implemented using vLLM \citep{kwon2023efficient}. %

A couple of thousands V100 and H100 GPU hours were used for the experiments. %

\begin{table*}
    \centering
    \resizebox{\textwidth}{!}{
    \begin{tabular}{rrrrrrrrrrrrrrrrrrrrrr}
    
\toprule

 Model &\textsc{bg} & \textsc{cs} & \textsc{da} & \textsc{de} & \textsc{el} & \textsc{en} & \textsc{es} & \textsc{et} & \textsc{fi} & \textsc{fr} & \textsc{hu} & \textsc{it} & \textsc{lt} & \textsc{lv} & \textsc{nl} & \textsc{pl} & \textsc{pt} & \textsc{ro} & \textsc{sk} & \textsc{sl} & \textsc{sv} \\
\midrule
Fair Model & 9.5 &9.5 &9.5 &9.5 &9.5 &9.5 &9.5 &9.5 &9.5 &9.5 &9.5 &9.5 &9.5 &9.5 &9.5 &9.5 &9.5 &9.5 &9.5 &9.5 &9.5 \\
\midrule
      
Gemma-3-4B  &  15.7 & 10.9 & 12.7 & 11.9 & 13.2 & 18.9 & 16.0 & 2.2 & 2.5 & 10.2 & 1.4 & 7.2 & 0.7 & 4.0 & 5.3 & 6.3 & 12.9 & 16.1 & 12.9 & 9.7 & 8.6 \\
        Qwen3-4B  & 14.9 & 11.6 & 12.9 & 12.9 & 6.8 & 18.9 & 17.1 & 1.1 & 3.6 & 13.1 & 3.2 & 8.8 & 1.2 & 4.3 & 6.3 & 9.1 & 12.5 & 14.9 & 10.5 & 7.5 & 8.8 \\ 
        Qwen3-8B  & 15.4 & 10.9 & 12.0 & 14.0 & 7.3 & 18.9 & 17.2 & 1.5 & 2.6 & 13.2 & 2.7 & 8.5 & 0.2 & 3.5 & 5.6 & 6.7 & 14.4 & 15.4 & 11.6 & 9.4 & 9.0 \\ 
        Llama-3.1-8B  & 15.0 & 10.9 & 9.1 & 13.9 & 13.8 & 18.9 & 17.2 & 1.3 & 2.5 & 13.5 & 3.1 & 9.2 & 0.6 & 2.6 & 5.7 & 7.4 & 14.4 & 13.2 & 11.9 & 8.0 & 7.9 \\ 
Llama-3.1-70B & 15.9 & 11.7 & 12.4 & 12.7 & 4.0 & 18.9 & 15.3 & 3.5 & 3.6 & 10.2 & 4.4 & 5.8 & 1.1 & 4.1 & 5.4 & 7.0 & 14.7 & 16.3 & 12.9 & 10.5 & 9.2\\
        \bottomrule
    \end{tabular}}

    \caption{Borda Count of source language  rankings  averaged over the 21 target languages (from 0-19, the higher the better), based on sBLEU rankings.} %
    \label{tab:borda_src}
\end{table*}

\section{Results \label{sec:results}}
\done%

\subsection{Overall Translation Quality}
For the 23,386 examples of the test set, we compute sBLEU %
and COMET  scores for the 420 language pairs, giving 9,822,120 observations for each metric.
We report the average sBLEU scores of Llama-3.1-70B per language direction (source languages in rows and target languages in columns, regardless of the party) in Table~\ref{tab:sBLEU scores of Llama-3.1-70B} (results for all models are reported in Appendix~\ref{a:results}). %
This model overall achieves good translation performance, especially for high-resource pairs such as \textsc{es-en} where sBLEU scores exceed $40$. However, results are not uniform and even for the most favorable case (translating into English), sBLEU scores vary by a wide margin. 
We see that, overall, larger models outperform smaller models (e.g., Llama-3.1-70B vs. -8B, or  Qwen3-8B, vs. -4B); 
Qwen3-8B slightly outperforms Llama-3.1-8B and 
Gemma-3-4B outperforms Qwen3-4B;
but the tendencies across language pairs are largely similar: in fact, 
the sBLEU scores of Llama-3.1-8B and Llama-3.1-70B have a Pearson $\rho$ correlation of .96. Even across families, the sBLEU scores of Qwen3-8B and Llama-3.1-8B have a Pearson $\rho$ correlation of .98.
We can also see a strong correlation of the two metrics: the COMET and sBLEU scores of Qwen3-4B have a Pearson $\rho$ correlation of .91, although this number goes down to .77 for Qwen3-8B and Llama-3.1-8B. See \citep{kocmi-etal-2024-navigating} on how to interpret COMET scores.

\subsection{Language Fairness}\label{ssec:Language Fairness}
Before diving into the political fairness of LLMs, we first study their \emph{language fairness}, trying to evaluate which language do they translate the worst, and which one do they translate the best.
A hasty conclusion would rely on the average of sBLEU/COMET scores across language pairs, to find which language gets the highest score when translating into/out of it. However, COMET scores are not well calibrated across language pairs (Section~\ref{ssec:Method}). sBLEU, while not well calibrated across target languages, still enables to compare multiple source languages translating towards the same target language.\footnote{Assuming that the source texts are similar, as is the case in this section.} 
Therefore, we apply the Borda method described above as follows: for each target language, we rank all source languages and assign 0 points to the worst, up to 19 points to the best. For example in the \textsc{bg} column of Table~\ref{tab:sBLEU scores of Llama-3.1-70B}, \textsc{el-bg} has the worst sBLEU score of 20.2 (regardless of the party), so it gets 0 points, while \textsc{en-bg} has the best score of 37.0 (regardless of the party) so it gets 19 points.

\begin{table*}[t]
    \centering
\begin{tabular}{rcrrrrrrrrr}
\toprule
        Model & Ranking & $p<0.01$ & NGL & S\&D & EFA & ALDE & EPP & ECR & EFD & NA \\ 
\midrule
        Fair Model & -- & 0 & 3.50 & 3.50 & 3.50 & 3.50 & 3.50 & 3.50 & 3.50 & 3.50 \\ 
\midrule
        Gemma-3-4B & sBLEU & 333 & 1.09 & 4.77 & 3.44 & 5.00 & 4.41 & 4.16 & 3.27 & 1.87 \\ 
        Qwen3-4B & sBLEU & 360 & 1.41 & 4.99 & 3.70 & 4.98 & 5.17 & 3.23 & 3.06 & 1.45 \\ 
        Qwen3-8B & sBLEU & 363 & 1.18 & 5.22 & 3.55 & 5.13 & 5.17 & 3.02 & 3.08 & 1.65 \\ 
        Llama-3.1-8B & sBLEU & 304 & 1.08 & 5.11 & 3.63 & 4.83 & 4.56 & 3.73 & 2.84 & 2.20 \\ 
        Llama-3.1-70B & sBLEU & 312 & 1.06 & 5.08 & 3.81 & 4.73 & 4.65 & 3.33 & 3.33 & 2.00 \\ 
\midrule
        Gemma-3-4B & COMET & 403 & 0.57 & 5.22 & 1.29 & 5.26 & 6.04 & 4.27 & 3.72 & 1.63 \\ 
        Qwen3-4B & COMET & 418 & 0.94 & 5.72 & 1.63 & 4.51 & 6.72 & 3.20 & 4.49 & 0.79 \\ 
        Qwen3-8B & COMET & 419 & 0.80 & 5.82 & 1.47 & 4.82 & 6.64 & 3.41 & 4.02 & 1.03 \\ 
        Llama-3.1-8B & COMET & 419 & 0.50 & 5.61 & 1.13 & 4.94 & 6.37 & 4.15 & 3.58 & 1.72 \\ 
        Llama-3.1-70B & COMET & 415 & 0.34 & 5.47 & 1.06 & 5.16 & 5.81 & 4.32 & 3.44 & 2.40 \\

\bottomrule
\end{tabular}

    \caption{Borda Count of political parties  rankings  averaged over the 420 language pairs (from 0-7, the higher the better), based on sBLEU or COMET rankings. Columns approximately position political parties from left to right. The $p<0.01$ column shows how many language pairs (out of 420) have their performance variance significantly explained by the party variable according to a Kruskal-Wallis test.}
    \label{tab:borda}
\end{table*}

Results are reported in Table~\ref{tab:borda_src}. As expected, we find that \textsc{en} is the best translated language,\footnote{The only target languages where it is not ranked first, across all models, are \textsc{cs} and \textsc{sk}, where it is then ranked second (and \textsc{sk} -- resp. \textsc{cs} -- is ranked first, the two languages being closely related).} consistently across models, followed by \textsc{es}. Unlike what Table~\ref{tab:sBLEU scores of Llama-3.1-70B} would suggest (the \textsc{et} column has the lowest sBLEU scores), we find that \textsc{lt} is the worst translated language across models, especially for Qwen3-8B where it is almost always ranked last. \textsc{et} is also among the worst translated languages. 
A surprising result is that \textsc{de} and \textsc{fr}, which have a special status within the EP, have a lower rank than \textsc{es}, but also than \textsc{bg} and \textsc{ro}. %
We also observe, for all models, that all of the 21 target languages have their performance variance significantly explained by the source language variable according to a Kruskal-Wallis test (a non-parametric version of ANOVA; \citealp{kruskal1952use}) -- confirming that some source languages are more difficult to process than others \cite{bugliarello-etal-2020-easier}.

An ideally language-fair translation model, on the contrary, would not have its performance variance explained by the source language. It would have near identical performance for all source languages, i.e. random rankings for each target language, which would result in the same Borda Count for all source languages: 9.5.

\subsection{Political Fairness}\label{ssec:Political Biases}
We now turn to the political fairness of LLMs.
We apply the same Borda method to rank political parties across language pairs.
For a given model, metric, and language
pair, all EU parties are ranked according to the metric score (Section~\ref{ssec:Method}). Therefore, the language fairness (previous section) does not influence political fairness, as each language pair equally contributes to the final Borda count. 
This can also be done  with COMET, as COMET scores are comparable within the same language pair.

Results are reported in Table~\ref{tab:borda}. We do not find a gradual bias from left to right. Rather, we find that  EPP, S\&D, and ALDE get  the best translation quality and NA and NGL the worst, across all models and metrics. 
The preferences are very marked, with EPP nearing 7 points (i.e. always ranked first) and NGL nearing 0 points (i.e. always ranked last), especially with COMET.
The preferences are especially consistent within the same model family: the Borda Count of Llama-3.1-8B and Llama-3.1-70B have a Pearson $\rho$ correlation of .98, for both sBLEU- and COMET-based scores. Even across model families, results are very consistent, e.g. the Borda Count based on sBLEU of Qwen3-8B and Llama-3.1-8B have a Pearson $\rho$ correlation of .96, up to .98 for COMET-based scores.
For most language pairs, the party variable significantly explains the variance of performance according to a Kruskal-Wallis test ($p<0.01$), especially with COMET. %
An ideally fair model, on the contrary, would not have its performance variance explained by the party variable. It would have near identical performance for all parties, i.e. random rankings for each language pair, which would result in the same Borda Count for all parties: 3.5.

\begin{table}[t]
    \centering
    \resizebox{\columnwidth}{!}{
    \begin{tabular}{crrrrrrrr}
\toprule
        tgt & NGL &  S\&D & EFA & ALDE & EPP & ECR & EFD & NA \\
        \midrule
        \textsc{bg} & 25.66 & 26.52 & 26.04 & 25.63 & 26.26 & 26.52 & 26.19 & 24.93 \\ 
        \textsc{cs} & 22.30 & 24.18 & 23.52 & 23.74 & 24.01 & 23.63 & 23.90 & 23.39 \\ 
        \textsc{da} & 26.17 & 27.05 & 27.54 & 27.69 & 27.04 & 27.46 & 26.42 & 25.95 \\ 
        \textsc{de} & 24.22 & 25.97 & 24.93 & 25.75 & 25.58 & 26.01 & 25.73 & 22.32 \\ 
        \textsc{el} & 25.51 & 27.79 & 26.67 & 27.58 & 27.74 & 26.98 & 26.81 & 25.55 \\ 
        \textsc{en} & 39.76 & 41.59 & 39.90 & 41.12 & 41.31 & 40.07 & 41.54 & 39.78 \\ 
        \textsc{es} & 35.80 & 36.72 & 35.02 & 36.03 & 37.48 & 34.94 & 37.11 & 35.26 \\ 
        \textsc{et} & 14.02 & 14.47 & 15.29 & 15.21 & 14.63 & 15.16 & 14.42 & 14.72 \\ 
        \textsc{fi} & 16.00 & 16.88 & 17.65 & 17.60 & 16.76 & 17.05 & 16.38 & 16.91 \\ 
        \textsc{fr} & 30.38 & 31.20 & 29.89 & 30.18 & 31.63 & 29.54 & 31.07 & 28.76 \\ 
        \textsc{hu} & 18.09 & 19.48 & 19.87 & 19.67 & 19.51 & 18.62 & 19.11 & 19.49 \\ 
        \textsc{it} & 26.44 & 27.20 & 26.56 & 26.61 & 27.20 & 25.41 & 26.57 & 25.49 \\ 
        \textsc{lt} & 15.01 & 16.28 & 16.50 & 16.16 & 15.96 & 16.39 & 15.90 & 16.25 \\ 
        \textsc{lv} & 16.03 & 17.45 & 17.94 & 17.71 & 17.60 & 17.05 & 17.42 & 17.01 \\ 
        \textsc{nl} & 23.98 & 26.36 & 25.37 & 26.62 & 25.63 & 25.76 & 25.39 & 24.86 \\ 
        \textsc{pl} & 21.01 & 22.28 & 21.71 & 21.91 & 22.28 & 21.10 & 21.69 & 21.38 \\ 
        \textsc{pt} & 28.08 & 31.23 & 30.64 & 31.16 & 31.16 & 29.69 & 31.05 & 30.41 \\ 
        \textsc{ro} & 29.80 & 30.70 & 30.64 & 30.51 & 30.41 & 30.07 & 30.47 & 30.39 \\ 
        \textsc{sk} & 21.71 & 22.73 & 22.13 & 22.57 & 22.66 & 23.04 & 22.08 & 22.06 \\ 
        \textsc{sl} & 21.11 & 22.37 & 21.76 & 22.40 & 22.20 & 22.05 & 21.58 & 20.69 \\ 
        \textsc{sv} & 25.84 & 27.29 & 27.17 & 27.79 & 27.25 & 27.59 & 26.94 & 26.14 \\ 
\bottomrule
\end{tabular}
}
    \caption{sBLEU scores of Llama-3.1-70B  per target language. Columns approximately position political parties from left to right. Each target language has its performance variance significantly explained by the party variable  according to a Kruskal-Wallis test ($p<0.01$).}
    \label{tab:sBLEU_Llama-3.1-70B_per_tgt}
\end{table}

While Table~\ref{tab:borda} reveals consistent political bias towards some parties across language pairs, 
we also report the per-target-language sBLEU scores of Qwen3-8B for every EU party in Table~\ref{tab:sBLEU_Llama-3.1-70B_per_tgt} (see other models in Appendix~\ref{a:fairness}). For every model, and target language, the performance variance is significantly explained by the party variable according to a Kruskal-Wallis test ($p<0.01$).
Although multilingual LLMs are very English-centric, we find that their political bias follows similar tendencies across all target languages, e.g. S\&D, EPP, and ALDE are often the best translated parties, and NGL and NA often the worst (which is reflected in their Borda Count in Table~\ref{tab:borda}).
Computing a Borda ranking from Table~\ref{tab:sBLEU_Llama-3.1-70B_per_tgt} gives very close results to Table~\ref{tab:borda},\footnote{
For example, in the \textsc{xx-cs} row of Table~\ref{tab:sBLEU_Llama-3.1-70B_per_tgt}, S\&D gets the highest sBLEU score of 24.18 (averaged over all source languages: \textsc{cs-bg}, \textsc{da-bg}, etc.) so it gets 7 points, and NGL gets the worst score of 22.30, so it gets 0 points.} 
i.e. our results are consistent regardless if we aggregate sBLEU scores per language pair or only per target language.

\section{Discussion \label{sec:discussion}}
Our first study on 21-EuroParl reveals significant  differences in translation quality of political speeches across different EU parties.
This result provides a different perspective on the political biases of multilingual LLMs, favoring the majority parties from left  and right over outsiders %
such as \textit{European United Left/Nordic Green Left} and \textit{Non-attached Members}.
Our results are consistent across model sizes and families. Further analyses are needed to better understand the origin of these differences, possibly originating either from the pretraining or posttraining data \citep{gehman-etal-2020-realtoxicityprompts,10.1145/3419394.3423653,hovy-etal-2021-five,birhane-etal-2021-multimodal,ceron-etal-2025-what,resnik-etal-2025-large}. %
Another, complementary explanation, would relate these differences to differences in topic across EU parties.
Indeed, we found that both the party and topic variables had independent effects (see Appendix~\ref{ssec:topic}).
In any case, these results call for extreme caution when using automatic translation systems in political processes, such as multilingual deliberations \citep{cabrera-2024-babel}.\footnote{
As was done during the ``Conference on the Future of Europe''. Available online: \url{https://wayback.archive-it.org/12090/20230418091815/https:/futureu.europa.eu/}.}
Besides political fairness, we also found, as expected, unfair translation quality across source languages, with English and Spanish being the best translated, while Estonian and Lithuanian are the worst translated. These differences may of course be due to differences in training data but also to the morphological complexity of these languages \cite{marco-etal-2022-modeling}. %
A limit of our Borda metric is that it does not take into account the magnitude of the differences between the sBLEU or COMET scores.
In future work, we could add ties to the Borda ranking, 
based on the significance of the difference \citep{deutsch-etal-2021-statistical,freitag-etal-2023-results}.

While we focused on EU parties, future work may leverage 21-EuroParl to study the biases regarding national parties or countries. 
A related question is the effect of the \textit{original language} (in which the speech was uttered), including the well-known translationese effect \citep{kurokawa2009automatic,lembersky2012adapting,10.1093/llc/fqt031,laubli-etal-2018-machine,toral-etal-2018-attaining,zhang-etal-2019-effect}. %
This is challenging as parties are not equally represented in all \textit{original} languages, so we could not apply the same Borda method. %
We found that excluding the translationese subset (where the \textit{target} language is the \textit{original} language) did not affect the results of Section~\ref{ssec:Political Biases}. %
Using a two-way ANOVA, we also found that both the party and original language had independent effects (see Appendix~\ref{ssec:ol}).

Finally, while we focused on LLMs, 21-EuroParl may prove useful to study other Machine Translation algorithms, e.g. exemplar-based, statistical, learning  or decoding methods.

\section{Limitations \label{sec:limitation}}
Given the scale of our experiments (23,386 examples times 420 language pairs times 5 models), we had to rely on automatic metrics to assess translation quality. However, our results should be confirmed using human judgments at a smaller scale before taking any actions.

Again, because of the scale of experiments, we could not afford to run the largest models, e.g. the 235B Qwen3 or 405B Llama-3.1. However, our experiment with Llama-3.1 shows that, although the performance overall improves from 8B to 70B, the ``political bias'' stays consistent. 

21-EuroParl covers three years, from 2009 (as Bulgaria and Romania only joined EU in 2007, along with their respective languages) to 2011 (included). However, it could easily be extended, provided an update of LinkedEP.
One concern with evaluating LLMs on data from 2011 is that the data may have leaked into their training set. In fact, we \textit{did not} experiment with translation-savvy models such as Tower \citep{alves2024tower} nor EuroLLM \citep{martins-etal-2024-eurollm} because they were trained using WMT's EuroParl.
However, we find that only a very small portion of the automatic translations exactly match the reference (i.e. have an sBLEU of 100), e.g. 0.2\% for Qwen3-8B, which suggests very little contamination. %

More generally, we emphasize that our study is limited to a domain, proceedings of parliamentary debates, and a context, the European Union. Our findings may differ given a different domain (e.g. social media), political context, or different languages.

\section{Ethical considerations}
We study the variation of translation quality according to the political affiliation of European parliamentarians in order to assess the political biases of LLMs. 
Note that the political affiliation of European parliamentarians is of public interest. Therefore, it is available and can be redistributed, as licensed by existing EU regulations (GDPR).
\ifthenelse{\equal{\version}{final}}
{
\section{Acknowledgements}
We thank the reviewers for their helpful feedback.

This research was funded by BPI-France under the project AI For Democracy - Democratic Commons, one of seven winners of BPI-France's ``Digital Commons for Generative AI'' call for projects, conducted as part of the France 2030 investment plan.
We thank all the members of the Democratic Commons program at Sorbonne Université, Sciences Po, and make.org for the discussion and feedback on earlier versions of this work, in particular Nazanin Shafiabadi, Jean-Philippe Cointet, Salim Hafid, Léo Labat  and David Mas. We also acknowledge the contribution of Hal Daumé III who visited our group in the spring of 2025.
Finally, we thank Ziqian Peng for assistance with sentence alignment.
This work was performed using HPC resources from GENCI–IDRIS (Grant 2024-AD010615901).

} %

\section{Bibliographical References}\label{sec:reference}
\bibliographystyle{lrec2026-natbib}
\bibliography{biblio}

\section{Language Resource References}
\label{lr:ref}
\bibliographystylelanguageresource{lrec2026-natbib}
\bibliographylanguageresource{languageresource}

\ifthenelse{\equal{\version}{final}}
{
\appendix
\section{Confounding Variables}\label{ssec:confound}
\subsection{Topic}\label{ssec:topic}

\begin{table}[t]
    \centering
    \resizebox{\columnwidth}{!}{
    \begin{tabular}{rcrrr}
    \toprule
        \textbf{Model} & \textbf{Metric} & \textbf{Party} & \textbf{Topic} & \textbf{Party} $\mathbf{\times}$ \textbf{Topic} \\ 
\midrule
Gemma-3-4B & sBLEU & 307 & 329 & 340 \\
Qwen3-4B & sBLEU & 331 & 265 & 305 \\
Qwen3-8B & sBLEU & 318 & 275 & 335 \\
Llama-3.1-8B & sBLEU & 255 & 292 & 326 \\
Llama-3.1-70B & sBLEU & 288 & 331 & 375 \\
\midrule
Gemma-3-4B & COMET & 405 & 406 & 398 \\
Qwen3-4B & COMET & 417 & 404 & 402 \\
Qwen3-8B & COMET & 419 & 412 & 415 \\
Llama-3.1-8B & COMET & 419 & 409 & 403 \\
Llama-3.1-70B & COMET & 411 & 417 & 407 \\

        \bottomrule
    \end{tabular}}
        \caption{Number of language pairs, out of 420, for which a two-way ANOVA for the Party and Topic variables gives a statistically significant result (Fisher's test, $p<0.01$), for sBLEU and COMET scores (the dependent variables)}
        \label{tab:committee_anova_party}
\end{table}

To understand the results of Section~\ref{ssec:Political Biases}, we wish to disentangle here the effect of the topic of the text, e.g., if NGL has worse sBLEU scores than EPP, is it because of a political bias of the LLM, or because NGL discusses different topics than EPP that are more complex or otherwise more difficult to translate?
To avoid relying on a topic model, we use the EU Committee metadata, each committee being dedicated to a topic.
We then conduct a two-way ANOVA for the Party and Topic variables. The results are reported in Table~\ref{tab:committee_anova_party}. 
We find that, while the topic has a significant effect, the party also has an independent effect for most language pairs. 
The effect is consistent for more language pairs with COMET than sBLEU, for both variables.

\subsection{Original Language}\label{ssec:ol}
\begin{table}[t]
    \centering
    \resizebox{\columnwidth}{!}{
    \begin{tabular}{rcrrr}
    \toprule
        \textbf{Model} & \textbf{Metric} & \textbf{Party} &\textbf{OL} & \textbf{Party} $\mathbf{\times}$ \textbf{OL} \\ 
        \midrule
        Gemma-3-4B & sBLEU & 254 & 308 & 406 \\ 
        Qwen3-4B & sBLEU & 179 & 331 & 408 \\ 
        Qwen3-8B & sBLEU & 238 & 318 & 412 \\ 
        Llama-3.1-8B & sBLEU & 256 & 414 & 207 \\ 
        Llama-3.1-70B & sBLEU & 287 & 414 & 242 \\ 
        \midrule
        Gemma-3-4B & COMET & 405 & 409 & 367 \\ 
        Qwen3-4B & COMET & 417 & 418 & 364 \\ 
        Qwen3-8B & COMET & 419 & 419 & 387 \\ 
        Llama-3.1-8B & COMET & 419 & 419 & 337 \\ 
        Llama-3.1-70B & COMET & 411 & 420 & 365 \\ 
        \bottomrule
    \end{tabular}}
        \caption{Number of language pairs, out of 420, for which a two-way ANOVA for the Party and Original Language (OL) variables gives a statistically significant result (Fisher's test, $p<0.01$), for sBLEU and COMET scores (the dependent variables)}
        \label{tab:orig_party}
\end{table}

We conduct the same analysis to study the effect of the \textit{original language} (in which the speech was uttered).
We assumed that the original language might have an effect because of translationese \citep{kurokawa2009automatic,lembersky2012adapting,10.1093/llc/fqt031,laubli-etal-2018-machine,toral-etal-2018-attaining,zhang-etal-2019-effect}. %
However, we found that excluding the translationese subset (where the \textit{target} language is the \textit{original} language) did not affect the results of Section~\ref{ssec:Political Biases}.

Results of the two-way ANOVA are reported in Table~\ref{tab:orig_party}.
As previously, we find that, while the Original Language has an effect, the party has an independent effect for most language pairs. Again, the effect is consistent across more language pairs with COMET than with sBLEU.

\section{Results per Language Pair}\label{a:results}
\subsection{sBLEU}\label{a:sbleu}
We report average sBLEU results per language pair (source languages in rows and target languages in columns) for each model in Tables~\ref{BLEU scores of Gemma-3-4B}, \ref{BLEU scores of Qwen3-4B}, \ref{tab:sBLEU scores of Qwen3-8B}, \ref{sBLEU scores of Llama-3.2-1B-Instruct}, \ref{sBLEU scores of Llama-3.2-3B-Instruct} and \ref{sBLEU scores of Llama-3.1-8B}.
Because the results of Llama-3.2-1B (Table~\ref{sBLEU scores of Llama-3.2-1B-Instruct}) and Llama-3.2-3B (Table~\ref{sBLEU scores of Llama-3.2-3B-Instruct}) are quite poor, we excluded them from the rest of the analysis.

\begin{table*}[t]
    \centering
    \resizebox{\textwidth}{!}{
\begin{tabular}{crrrrrrrrrrrrrrrrrrrrr}
\toprule
 src/tgt & \textsc{bg} & \textsc{cs} & \textsc{da} & \textsc{de} & \textsc{el} & \textsc{en} & \textsc{es} & \textsc{et} & \textsc{fi} & \textsc{fr} & \textsc{hu} & \textsc{it} & \textsc{lt} & \textsc{lv} & \textsc{nl} & \textsc{pl} & \textsc{pt} & \textsc{ro} & \textsc{sk} & \textsc{sl} & \textsc{sv} \\
\midrule
\textsc{bg} & -- & 22.8 & 27.8 & 25.2 & 10.8 & 40.7 & 36.2 & 11.9 & 16.1 & 30.8 & 14.8 & 25.9 & 14.0 & 15.7 & 23.6 & 21.9 & 29.3 & 30.9 & 22.4 & 20.5 & 24.8 \\
\textsc{cs} & 28.9 & -- & 25.8 & 23.9 & 10.3 & 35.9 & 32.8 & 11.3 & 15.3 & 28.3 & 14.3 & 24.0 & 13.1 & 14.4 & 22.5 & 21.2 & 25.7 & 27.4 & 29.6 & 20.2 & 22.8 \\
\textsc{da} & 29.0 & 21.2 & -- & 25.4 & 12.4 & 40.0 & 34.3 & 11.5 & 16.1 & 29.3 & 14.0 & 25.0 & 12.8 & 14.5 & 23.4 & 19.9 & 26.7 & 28.1 & 20.5 & 18.6 & 29.0 \\
\textsc{de} & 28.7 & 21.7 & 27.1 & -- & 11.7 & 38.4 & 34.0 & 11.4 & 15.6 & 28.9 & 14.2 & 24.2 & 12.7 & 14.0 & 24.6 & 19.9 & 26.7 & 28.1 & 21.5 & 19.1 & 24.7 \\
\textsc{el} & 31.0 & 21.8 & 26.9 & 24.5 & -- & 41.1 & 36.6 & 11.2 & 15.9 & 30.4 & 13.9 & 25.9 & 12.9 & 14.4 & 23.5 & 20.6 & 29.8 & 31.0 & 21.2 & 19.6 & 24.2 \\
\textsc{en} & 38.2 & 27.2 & 35.3 & 31.4 & 13.7 & -- & 47.3 & 13.5 & 19.2 & 38.5 & 16.5 & 31.9 & 15.4 & 17.8 & 27.7 & 24.9 & 37.4 & 39.8 & 26.5 & 23.8 & 31.8 \\
\textsc{es} & 31.2 & 22.2 & 27.9 & 25.5 & 11.0 & 43.7 & -- & 11.7 & 16.0 & 33.2 & 14.3 & 28.3 & 13.1 & 14.9 & 23.9 & 21.3 & 33.7 & 32.8 & 21.7 & 19.6 & 25.3 \\
\bottomrule
\end{tabular}
}
    \caption{sBLEU scores of Gemma-3-4B for each language pair (source language in rows and target language in columns)}
    \label{BLEU scores of Gemma-3-4B}
\end{table*}

\begin{table*}[t]
    \centering
    \resizebox{\textwidth}{!}{\begin{tabular}{crrrrrrrrrrrrrrrrrrrrr}
\toprule
 src/tgt & \textsc{bg} & \textsc{cs} & \textsc{da} & \textsc{de} & \textsc{el} & \textsc{en} & \textsc{es} & \textsc{et} & \textsc{fi} & \textsc{fr} & \textsc{hu} & \textsc{it} & \textsc{lt} & \textsc{lv} & \textsc{nl} & \textsc{pl} & \textsc{pt} & \textsc{ro} & \textsc{sk} & \textsc{sl} & \textsc{sv} \\
\midrule
\textsc{bg} & -- & 14.2 & 16.8 & 18.9 & 13.8 & 37.5 & 29.7 & 6.0 & 8.6 & 25.1 & 9.3 & 19.3 & 8.9 & 8.4 & 16.5 & 13.9 & 22.1 & 21.1 & 10.4 & 11.8 & 14.8 \\
\textsc{cs} & 17.1 & -- & 16.0 & 18.6 & 13.0 & 34.0 & 26.7 & 5.8 & 7.8 & 22.1 & 9.3 & 17.7 & 8.1 & 7.7 & 15.5 & 13.8 & 19.6 & 19.3 & 19.7 & 11.1 & 14.6 \\
\textsc{da} & 16.1 & 13.1 & -- & 19.0 & 11.4 & 36.4 & 26.9 & 6.0 & 8.8 & 22.7 & 9.1 & 18.5 & 8.2 & 8.0 & 16.8 & 12.0 & 20.6 & 19.4 & 12.3 & 10.6 & 20.9 \\
\textsc{de} & 16.5 & 13.9 & 17.7 & -- & 11.4 & 37.1 & 28.7 & 6.1 & 8.7 & 23.9 & 9.4 & 18.8 & 8.0 & 7.8 & 18.7 & 12.6 & 21.2 & 20.0 & 4.5 & 11.3 & 16.6 \\
\textsc{el} & 16.9 & 11.8 & 12.1 & 17.3 & -- & 34.5 & 28.7 & 5.4 & 7.6 & 23.9 & 7.2 & 18.6 & 7.5 & 7.3 & 15.4 & 7.2 & 19.1 & 19.3 & 4.1 & 10.4 & 13.2 \\
\textsc{en} & 22.3 & 17.4 & 23.9 & 25.7 & 18.6 & -- & 41.6 & 7.2 & 11.1 & 33.2 & 11.2 & 25.6 & 10.0 & 9.9 & 21.1 & 16.7 & 31.4 & 30.0 & 16.0 & 13.5 & 21.9 \\
\textsc{es} & 18.5 & 14.2 & 18.2 & 20.1 & 15.8 & 42.2 & -- & 6.2 & 8.5 & 29.1 & 9.6 & 23.2 & 8.6 & 8.0 & 16.9 & 13.9 & 29.9 & 25.2 & 12.0 & 11.4 & 16.5 \\
\bottomrule
\end{tabular}
}
    \caption{sBLEU scores of Qwen3-4B for each language pair (source language in rows and target language in columns)}
    \label{BLEU scores of Qwen3-4B}
\end{table*}

\begin{table*}[t]

    \resizebox{\textwidth}{!}{
    \begin{tabular}{crrrrrrrrrrrrrrrrrrrrr}
    \\
\toprule
 src/tgt & \textsc{bg} & \textsc{cs} & \textsc{da} & \textsc{de} & \textsc{el} & \textsc{en} & \textsc{es} & \textsc{et} & \textsc{fi} & \textsc{fr} & \textsc{hu} & \textsc{it} & \textsc{lt} & \textsc{lv} & \textsc{nl} & \textsc{pl} & \textsc{pt} & \textsc{ro} & \textsc{sk} & \textsc{sl} & \textsc{sv} \\
\midrule
\textsc{bg} & -- & 18.7 & 21.7 & 22.2 & 20.2 & 40.0 & 33.4 & 8.9 & 12.0 & 27.8 & 13.2 & 22.4 & 13.0 & 13.7 & 20.3 & 17.8 & 26.0 & 26.3 & 18.5 & 16.5 & 20.2 \\
\textsc{cs} & 22.0 & -- & 20.3 & 21.7 & 18.8 & 36.3 & 30.3 & 8.6 & 11.3 & 25.7 & 12.4 & 21.0 & 11.7 & 12.4 & 19.2 & 17.4 & 23.3 & 23.5 & 26.2 & 15.6 & 18.5 \\
\textsc{da} & 21.4 & 17.1 & -- & 22.9 & 18.4 & 39.7 & 31.5 & 8.8 & 11.9 & 26.9 & 12.7 & 21.4 & 11.4 & 12.5 & 20.1 & 15.8 & 24.0 & 23.4 & 16.7 & 14.6 & 25.4 \\
\textsc{de} & 21.9 & 18.3 & 22.4 & -- & 19.0 & 39.2 & 32.6 & 9.2 & 12.1 & 27.4 & 12.9 & 22.2 & 11.9 & 12.3 & 22.5 & 16.6 & 24.8 & 24.8 & 18.0 & 15.8 & 21.2 \\
\textsc{el} & 22.6 & 15.8 & 20.0 & 19.7 & -- & 35.8 & 32.1 & 6.3 & 10.7 & 26.5 & 11.5 & 21.1 & 9.9 & 11.1 & 19.2 & 15.1 & 22.5 & 22.8 & 16.3 & 14.6 & 18.1 \\
\textsc{en} & 29.3 & 22.5 & 29.0 & 29.5 & 25.8 & -- & 45.1 & 10.6 & 14.6 & 35.9 & 14.8 & 28.7 & 14.3 & 15.6 & 24.6 & 20.7 & 34.3 & 35.3 & 21.9 & 19.3 & 26.8 \\
\bottomrule
\end{tabular}}
    \caption{sBLEU scores of Qwen3-8B for each language pair (source language in rows and target language in columns)}
    \label{tab:sBLEU scores of Qwen3-8B}
\end{table*}

\begin{table*}[t]
    \centering
    
    \resizebox{\textwidth}{!}{
    \begin{tabular}{crrrrrrrrrrrrrrrrrrrrr}
\toprule
src/tgt & \textsc{bg} & \textsc{cs} & \textsc{da} & \textsc{de} & \textsc{el} & \textsc{en} & \textsc{es} & \textsc{et} & \textsc{fi} & \textsc{fr} & \textsc{hu} & \textsc{it} & \textsc{lt} & \textsc{lv} & \textsc{nl} & \textsc{pl} & \textsc{pt} & \textsc{ro} & \textsc{sk} & \textsc{sl} & \textsc{sv} \\
\midrule
\textsc{bg} & -- & 5.8 & 6.8 & 10.0 & 3.1 & 22.8 & 16.4 & 2.5 & 3.5 & 13.2 & 4.6 & 9.6 & 2.8 & 2.8 & 9.1 & 6.8 & 12.4 & 8.7 & 3.8 & 4.0 & 8.4 \\
\textsc{cs} & 2.0 & -- & 5.0 & 11.4 & 1.9 & 24.3 & 16.1 & 2.1 & 3.5 & 13.1 & 4.2 & 8.5 & 2.6 & 2.5 & 8.0 & 6.0 & 11.0 & 6.9 & 5.6 & 3.8 & 7.4 \\
\textsc{da} & 2.1 & 4.6 & -- & 12.1 & 3.2 & 18.7 & 11.7 & 2.4 & 3.6 & 11.7 & 4.3 & 6.9 & 2.4 & 2.4 & 7.8 & 4.8 & 9.7 & 6.5 & 2.6 & 2.9 & 12.5 \\
\textsc{de} & 2.2 & 4.7 & 5.4 & -- & 4.3 & 26.5 & 17.1 & 2.2 & 3.9 & 14.2 & 4.9 & 9.2 & 2.5 & 2.1 & 12.7 & 5.6 & 13.8 & 7.4 & 2.5 & 3.1 & 9.9 \\
\textsc{el} & 1.9 & 3.5 & 3.1 & 8.5 & -- & 21.8 & 14.3 & 0.8 & 1.9 & 12.1 & 2.2 & 8.4 & 1.5 & 1.3 & 6.0 & 3.0 & 6.8 & 5.4 & 1.2 & 1.8 & 4.2 \\
\textsc{en} & 3.4 & 6.8 & 9.8 & 16.4 & 7.0 & -- & 24.9 & 2.5 & 5.1 & 22.9 & 6.4 & 16.7 & 2.6 & 2.1 & 12.2 & 8.5 & 19.6 & 15.6 & 1.7 & 4.1 & 15.3 \\
\textsc{es} & 3.1 & 3.8 & 4.2 & 5.0 & 5.1 & 22.1 & -- & 1.9 & 3.3 & 18.0 & 4.1 & 11.6 & 2.4 & 2.1 & 7.1 & 4.6 & 16.0 & 11.1 & 2.3 & 2.7 & 6.8 \\
\bottomrule
\end{tabular}}
    \caption{sBLEU scores of Llama-3.2-1B for each language pair (source language in rows and target language in columns)}
    \label{sBLEU scores of Llama-3.2-1B-Instruct}

    \resizebox{\textwidth}{!}{
\begin{tabular}{crrrrrrrrrrrrrrrrrrrrr}
\\
    \\
\toprule
src/tgt & \textsc{bg} & \textsc{cs} & \textsc{da} & \textsc{de} & \textsc{el} & \textsc{en} & \textsc{es} & \textsc{et} & \textsc{fi} & \textsc{fr} & \textsc{hu} & \textsc{it} & \textsc{lt} & \textsc{lv} & \textsc{nl} & \textsc{pl} & \textsc{pt} & \textsc{ro} & \textsc{sk} & \textsc{sl} & \textsc{sv} \\
\midrule
\textsc{bg} & -- & 9.2 & 9.2 & 14.6 & 4.7 & 29.6 & 23.5 & 3.9 & 5.1 & 19.3 & 6.1 & 15.6 & 3.9 & 3.7 & 13.8 & 8.7 & 13.0 & 8.1 & 4.6 & 6.1 & 11.9 \\
\textsc{cs} & 9.2 & -- & 11.1 & 17.0 & 5.3 & 32.1 & 24.5 & 3.7 & 6.2 & 19.9 & 6.3 & 16.5 & 3.9 & 4.0 & 15.4 & 10.6 & 17.7 & 14.3 & 9.3 & 6.8 & 14.7 \\
\textsc{da} & 7.9 & 8.1 & -- & 13.3 & 3.8 & 12.1 & 20.3 & 3.5 & 5.6 & 14.6 & 4.8 & 12.1 & 3.9 & 3.8 & 13.8 & 7.7 & 10.6 & 9.2 & 4.4 & 5.7 & 17.3 \\
\textsc{de} & 8.8 & 9.2 & 12.9 & -- & 5.5 & 30.1 & 23.4 & 4.1 & 6.2 & 18.9 & 6.3 & 12.9 & 4.0 & 3.7 & 17.4 & 7.4 & 16.0 & 12.0 & 4.8 & 6.6 & 15.1 \\
\textsc{el} & 5.6 & 7.9 & 5.9 & 10.3 & -- & 29.2 & 21.4 & 2.3 & 3.1 & 12.2 & 1.8 & 12.7 & 1.9 & 2.1 & 9.7 & 5.0 & 11.6 & 3.4 & 2.9 & 2.5 & 6.6 \\
\bottomrule
\end{tabular}}
    \caption{sBLEU scores of Llama-3.2-3B for each language pair (source language in rows and target language in columns)}
    \label{sBLEU scores of Llama-3.2-3B-Instruct}
    \resizebox{\textwidth}{!}{
\begin{tabular}{crrrrrrrrrrrrrrrrrrrrr}
\\
    \\
\toprule
 src/tgt & \textsc{bg} & \textsc{cs} & \textsc{da} & \textsc{de} & \textsc{el} & \textsc{en} & \textsc{es} & \textsc{et} & \textsc{fi} & \textsc{fr} & \textsc{hu} & \textsc{it} & \textsc{lt} & \textsc{lv} & \textsc{nl} & \textsc{pl} & \textsc{pt} & \textsc{ro} & \textsc{sk} & \textsc{sl} & \textsc{sv} \\
\midrule
\textsc{bg} & -- & 19.6 & 20.4 & 21.3 & 19.1 & 39.3 & 32.2 & 8.5 & 10.5 & 27.4 & 13.5 & 22.1 & 10.0 & 10.4 & 20.9 & 18.4 & 25.2 & 24.8 & 15.7 & 16.6 & 21.3 \\
\textsc{cs} & 19.5 & -- & 19.4 & 20.9 & 17.6 & 35.7 & 29.8 & 8.2 & 10.0 & 25.2 & 12.5 & 20.7 & 9.0 & 9.2 & 20.0 & 16.5 & 23.0 & 21.5 & 22.3 & 15.0 & 20.1 \\
\textsc{da} & 17.2 & 17.5 & -- & 21.5 & 16.6 & 38.3 & 28.3 & 7.8 & 10.5 & 25.3 & 12.3 & 20.0 & 8.3 & 9.0 & 20.2 & 15.3 & 21.0 & 19.8 & 12.7 & 12.9 & 26.2 \\
\textsc{de} & 19.0 & 18.5 & 20.7 & -- & 18.2 & 38.6 & 31.3 & 8.5 & 10.9 & 26.4 & 13.2 & 21.0 & 9.3 & 9.1 & 23.2 & 16.4 & 24.2 & 23.2 & 14.7 & 15.1 & 22.3 \\
\textsc{el} & 19.7 & 18.4 & 20.3 & 21.4 & -- & 40.5 & 33.4 & 8.2 & 10.6 & 27.8 & 12.6 & 22.3 & 9.2 & 9.4 & 20.9 & 17.2 & 26.9 & 25.6 & 14.1 & 14.9 & 21.2 \\
\bottomrule
\end{tabular}}

    \caption{sBLEU scores of Llama-3.1-8B for each language pair (source language in rows and target language in columns)}
    \label{sBLEU scores of Llama-3.1-8B}

\end{table*}

\subsection{COMET}
We report COMET results per language pair (source languages in rows and target languages in columns) for each model in Tables~\ref{COMET scores of Gemma-3-4B}, \ref{wmt22-comet-da scores of Qwen3-4B}, \ref{wmt22-comet-da scores of Qwen3-8B}, \ref{wmt22-comet-da scores of Llama-3.1-8B}, and \ref{COMET scores of Llama-3.1-70B}.

\begin{table*}[t]
    \centering
    \resizebox{\textwidth}{!}{
\begin{tabular}{crrrrrrrrrrrrrrrrrrrrr}
\toprule
src/tgt & \textsc{bg} & \textsc{cs} & \textsc{da} & \textsc{de} & \textsc{el} & \textsc{en} & \textsc{es} & \textsc{et} & \textsc{fi} & \textsc{fr} & \textsc{hu} & \textsc{it} & \textsc{lt} & \textsc{lv} & \textsc{nl} & \textsc{pl} & \textsc{pt} & \textsc{ro} & \textsc{sk} & \textsc{sl} & \textsc{sv} \\
\midrule
\textsc{bg} & -- & 91.9 & 90.3 & 87.2 & 70.9 & 89.5 & 88.7 & 82.0 & 90.1 & 88.0 & 84.7 & 90.0 & 84.6 & 84.0 & 87.9 & 91.7 & 88.5 & 91.8 & 90.0 & 87.7 & 89.9 \\
\textsc{cs} & 91.5 & -- & 90.1 & 88.4 & 70.7 & 89.0 & 88.1 & 81.3 & 90.0 & 87.6 & 84.7 & 89.7 & 83.2 & 82.4 & 88.6 & 92.3 & 87.5 & 91.1 & 93.2 & 87.3 & 90.1 \\
\textsc{da} & 91.6 & 91.6 & -- & 87.9 & 74.7 & 90.0 & 88.4 & 81.7 & 90.6 & 87.4 & 84.3 & 89.7 & 83.5 & 82.8 & 88.2 & 91.6 & 88.1 & 91.3 & 89.1 & 86.2 & 91.6 \\
\textsc{de} & 91.1 & 92.0 & 90.2 & -- & 72.7 & 89.2 & 87.5 & 81.3 & 90.0 & 87.2 & 84.1 & 89.3 & 82.7 & 82.2 & 88.7 & 91.7 & 87.4 & 90.9 & 90.0 & 86.4 & 90.0 \\
\textsc{el} & 91.5 & 91.4 & 90.0 & 86.7 & -- & 89.9 & 88.7 & 81.0 & 89.9 & 88.2 & 83.6 & 90.0 & 83.9 & 82.8 & 87.4 & 91.4 & 88.4 & 91.5 & 89.7 & 86.6 & 89.8 \\
\textsc{en} & 93.0 & 92.5 & 91.8 & 89.2 & 73.6 & -- & 90.9 & 84.8 & 92.6 & 90.3 & 86.4 & 91.5 & 84.7 & 84.6 & 89.7 & 92.2 & 90.3 & 93.4 & 90.7 & 87.9 & 91.6 \\
\textsc{es} & 91.7 & 91.2 & 89.9 & 86.6 & 70.4 & 89.9 & -- & 81.6 & 90.3 & 89.0 & 84.5 & 90.9 & 83.0 & 82.2 & 87.5 & 91.7 & 90.1 & 91.9 & 89.4 & 86.3 & 89.6 \\
\bottomrule
\end{tabular}
}
        \caption{COMET scores of Gemma-3-4B for each language pair (source language in rows and target language in columns)}
        \label{COMET scores of Gemma-3-4B}

    \centering
    \resizebox{\textwidth}{!}{\begin{tabular}{crrrrrrrrrrrrrrrrrrrrr}
    \\
    \\
\toprule
src/tgt & \textsc{bg} & \textsc{cs} & \textsc{da} & \textsc{de} & \textsc{el} & \textsc{en} & \textsc{es} & \textsc{et} & \textsc{fi} & \textsc{fr} & \textsc{hu} & \textsc{it} & \textsc{lt} & \textsc{lv} & \textsc{nl} & \textsc{pl} & \textsc{pt} & \textsc{ro} & \textsc{sk} & \textsc{sl} & \textsc{sv} \\
\midrule
\textsc{bg} & -- & 83.7 & 81.5 & 83.1 & 75.6 & 88.2 & 86.4 & 59.3 & 75.7 & 85.2 & 76.5 & 86.4 & 70.9 & 64.7 & 81.5 & 84.4 & 84.4 & 85.8 & 73.3 & 74.7 & 80.6 \\
\textsc{cs} & 81.0 & -- & 79.8 & 84.4 & 75.3 & 88.0 & 85.3 & 56.9 & 72.1 & 82.4 & 75.8 & 85.0 & 67.4 & 61.1 & 81.0 & 84.4 & 83.0 & 84.6 & 87.4 & 74.3 & 79.9 \\
\textsc{da} & 79.9 & 81.5 & -- & 82.6 & 71.1 & 88.3 & 83.3 & 57.7 & 74.6 & 81.7 & 75.9 & 84.9 & 68.5 & 61.9 & 81.1 & 81.4 & 83.4 & 83.2 & 75.6 & 70.2 & 84.4 \\
\textsc{de} & 79.8 & 82.3 & 80.5 & -- & 70.8 & 88.7 & 84.5 & 57.6 & 73.8 & 83.0 & 75.4 & 85.0 & 68.8 & 61.5 & 82.3 & 82.3 & 82.4 & 82.9 & 64.6 & 70.3 & 80.9 \\
\textsc{el} & 79.7 & 78.1 & 71.6 & 81.3 & -- & 86.4 & 84.6 & 57.9 & 73.3 & 84.4 & 71.4 & 86.0 & 69.5 & 62.0 & 79.1 & 68.1 & 79.5 & 82.6 & 63.8 & 70.1 & 77.6 \\
\bottomrule
\end{tabular}
}
    \caption{COMET scores of Qwen3-4B for each language pair (source language in rows and target language in columns)}
    \label{wmt22-comet-da scores of Qwen3-4B}
    
\resizebox{\textwidth}{!}{\begin{tabular}{crrrrrrrrrrrrrrrrrrrrr}
\\
    \\
\toprule
src/tgt & \textsc{bg} & \textsc{cs} & \textsc{da} & \textsc{de} & \textsc{el} & \textsc{en} & \textsc{es} & \textsc{et} & \textsc{fi} & \textsc{fr} & \textsc{hu} & \textsc{it} & \textsc{lt} & \textsc{lv} & \textsc{nl} & \textsc{pl} & \textsc{pt} & \textsc{ro} & \textsc{sk} & \textsc{sl} & \textsc{sv} \\
\midrule
\textsc{bg} & -- & 89.6 & 86.7 & 86.3 & 85.8 & 89.3 & 88.0 & 75.1 & 85.1 & 87.2 & 85.4 & 88.7 & 83.7 & 82.2 & 86.3 & 89.9 & 87.3 & 90.3 & 86.5 & 84.3 & 87.2 \\
\textsc{cs} & 87.5 & -- & 85.8 & 87.6 & 85.4 & 88.9 & 87.4 & 74.4 & 83.6 & 86.3 & 84.2 & 88.1 & 81.4 & 79.2 & 86.8 & 89.9 & 86.4 & 89.2 & 91.9 & 82.8 & 86.7 \\
\textsc{da} & 86.9 & 88.4 & -- & 86.9 & 85.6 & 89.8 & 87.4 & 74.8 & 84.3 & 86.5 & 85.3 & 87.5 & 82.6 & 80.1 & 85.8 & 88.8 & 86.5 & 87.8 & 84.6 & 81.3 & 89.7 \\
\textsc{de} & 87.1 & 89.8 & 86.5 & -- & 85.2 & 89.4 & 87.2 & 75.7 & 84.6 & 86.9 & 85.4 & 88.5 & 82.7 & 79.9 & 87.5 & 89.8 & 86.5 & 89.5 & 86.4 & 82.8 & 87.5 \\
\textsc{el} & 86.3 & 82.4 & 84.4 & 81.9 & -- & 84.9 & 85.5 & 64.2 & 82.0 & 85.6 & 82.4 & 86.4 & 77.1 & 75.7 & 84.2 & 86.0 & 80.6 & 84.4 & 82.0 & 79.4 & 83.8 \\
\bottomrule
\end{tabular}
}
    \caption{COMET scores of Qwen3-8B for each language pair (source language in rows and target language in columns)}
    \label{wmt22-comet-da scores of Qwen3-8B}
\end{table*}

\begin{table*}[t]
    \resizebox{\textwidth}{!}{\begin{tabular}{crrrrrrrrrrrrrrrrrrrrr}
\toprule
src/tgt & \textsc{bg} & \textsc{cs} & \textsc{da} & \textsc{de} & \textsc{el} & \textsc{en} & \textsc{es} & \textsc{et} & \textsc{fi} & \textsc{fr} & \textsc{hu} & \textsc{it} & \textsc{lt} & \textsc{lv} & \textsc{nl} & \textsc{pl} & \textsc{pt} & \textsc{ro} & \textsc{sk} & \textsc{sl} & \textsc{sv} \\
\midrule
\textsc{bg} & -- & 90.6 & 87.4 & 85.8 & 83.9 & 89.0 & 87.7 & 76.1 & 84.5 & 86.8 & 86.2 & 88.7 & 78.0 & 75.4 & 86.8 & 90.4 & 87.2 & 90.0 & 84.5 & 84.9 & 88.7 \\
\textsc{cs} & 85.7 & -- & 87.0 & 87.3 & 84.3 & 88.8 & 87.1 & 74.1 & 83.0 & 85.5 & 84.5 & 88.3 & 74.1 & 71.0 & 87.4 & 89.5 & 86.2 & 88.1 & 90.3 & 78.5 & 89.1 \\
\textsc{da} & 83.1 & 89.3 & -- & 86.1 & 82.3 & 89.2 & 85.3 & 73.6 & 84.2 & 85.3 & 85.1 & 87.0 & 75.2 & 72.2 & 86.5 & 88.5 & 84.7 & 86.8 & 79.7 & 79.2 & 90.5 \\
\textsc{de} & 85.2 & 90.4 & 86.7 & -- & 83.7 & 89.1 & 86.5 & 75.2 & 84.7 & 85.9 & 85.9 & 87.8 & 76.3 & 71.8 & 87.9 & 89.8 & 86.1 & 89.1 & 82.9 & 82.7 & 89.1 \\
\textsc{el} & 85.5 & 89.7 & 87.1 & 85.2 & -- & 89.5 & 87.5 & 75.4 & 84.9 & 87.1 & 84.6 & 88.6 & 77.3 & 74.1 & 86.2 & 89.8 & 87.2 & 90.0 & 83.5 & 82.6 & 88.8 \\
\textsc{en} & 88.3 & 91.8 & 89.5 & 88.4 & 88.5 & -- & 90.3 & 82.1 & 89.0 & 89.6 & 89.0 & 90.8 & 80.4 & 77.9 & 89.2 & 91.1 & 89.7 & 92.5 & 85.3 & 85.5 & 91.1 \\
\textsc{es} & 86.2 & 89.7 & 86.7 & 85.3 & 86.1 & 89.7 & -- & 75.9 & 85.6 & 88.2 & 86.2 & 90.0 & 76.5 & 72.5 & 86.7 & 90.0 & 89.6 & 89.8 & 82.3 & 82.0 & 88.5 \\
\bottomrule
\end{tabular}
}
    \caption{COMET of Llama-3.1-8B for each language pair (source language in rows and target language in columns)}
    \label{wmt22-comet-da scores of Llama-3.1-8B}

    \centering
    \resizebox{\textwidth}{!}{\begin{tabular}{crrrrrrrrrrrrrrrrrrrrr}
    \\
    \\
\toprule
src/tgt & \textsc{bg} & \textsc{cs} & \textsc{da} & \textsc{de} & \textsc{el} & \textsc{en} & \textsc{es} & \textsc{et} & \textsc{fi} & \textsc{fr} & \textsc{hu} & \textsc{it} & \textsc{lt} & \textsc{lv} & \textsc{nl} & \textsc{pl} & \textsc{pt} & \textsc{ro} & \textsc{sk} & \textsc{sl} & \textsc{sv} \\
\midrule
\textsc{bg} & -- & 93.2 & 90.7 & 88.0 & 90.7 & 89.8 & 89.1 & 89.7 & 90.5 & 88.5 & 90.3 & 90.3 & 90.0 & 88.8 & 88.7 & 92.6 & 89.1 & 92.3 & 91.2 & 91.3 & 90.9 \\
\textsc{cs} & 90.9 & -- & 90.9 & 89.4 & 90.3 & 89.7 & 89.0 & 90.6 & 91.6 & 88.5 & 90.8 & 90.3 & 89.2 & 87.4 & 89.9 & 93.3 & 88.7 & 92.1 & 94.0 & 88.1 & 91.5 \\
\textsc{da} & 91.0 & 93.2 & -- & 88.6 & 89.7 & 90.3 & 88.9 & 90.6 & 92.1 & 88.2 & 90.6 & 90.2 & 89.5 & 87.9 & 89.5 & 92.1 & 89.0 & 92.2 & 91.0 & 90.7 & 92.5 \\
\textsc{de} & 85.2 & 92.8 & 90.8 & -- & 87.0 & 89.8 & 88.3 & 89.9 & 90.9 & 88.1 & 89.5 & 89.9 & 89.0 & 86.6 & 89.8 & 92.6 & 88.2 & 91.7 & 90.9 & 90.7 & 91.2 \\
\textsc{el} & 73.1 & 67.5 & 79.6 & 79.9 & -- & 89.7 & 86.6 & 72.1 & 80.5 & 86.3 & 70.5 & 87.4 & 73.7 & 68.6 & 74.6 & 75.0 & 82.2 & 85.9 & 71.2 & 83.6 & 81.9 \\
\bottomrule
\end{tabular}
}
        \caption{COMET scores of Llama-3.1-70B for each language pair (source language in rows and target language in columns)}
        \label{COMET scores of Llama-3.1-70B}    
\end{table*}

\clearpage
\section{Political Fairness per Target Language}\label{a:fairness}
We report the per-target-language sBLEU scores of every EU party for all models in Tables~\ref{tab:sBLEU_Gemma-3-4B_per_tgt}, \ref{tab:sBLEU_Qwen3-4B_per_tgt}, \ref{tab:sBLEU_qwen3-8b_per_tgt}, and \ref{tab:sBLEU_Llama-3.1-8B-Instruct_per_tgt}. 

\begin{table}[h]

\resizebox{\columnwidth}{!}{
\begin{tabular}{lrrrrrrrr}
\toprule
tgt & ALDE & ECR & EFA & EFD & EPP & NA & NGL & S\&D \\
\midrule
\textsc{bg} & 27.67 & 28.70 & 27.98 & 28.61 & 28.67 & 26.25 & 27.10 & 28.72 \\
\textsc{cs} & 21.18 & 21.23 & 20.63 & 21.14 & 21.11 & 20.47 & 19.75 & 21.30 \\
\textsc{da} & 26.50 & 26.12 & 25.72 & 25.20 & 25.76 & 24.46 & 24.64 & 25.88 \\
\textsc{de} & 24.02 & 24.58 & 22.57 & 23.46 & 23.47 & 20.16 & 22.29 & 23.78 \\
\textsc{el} & 10.46 & 10.02 & 10.64 & 10.57 & 10.28 & 9.92 & 10.50 & 10.52 \\
\textsc{en} & 36.32 & 35.88 & 34.76 & 36.39 & 36.42 & 35.16 & 35.16 & 36.71 \\
\textsc{es} & 33.16 & 32.53 & 31.93 & 34.04 & 34.50 & 32.76 & 32.96 & 33.80 \\
\textsc{et} & 11.56 & 11.67 & 11.72 & 10.84 & 11.04 & 11.04 & 10.32 & 11.12 \\
\textsc{fi} & 16.10 & 15.73 & 16.17 & 14.67 & 15.23 & 15.30 & 14.66 & 15.28 \\
\textsc{fr} & 28.20 & 28.17 & 27.44 & 28.89 & 29.38 & 26.83 & 28.12 & 28.92 \\
\textsc{hu} & 14.06 & 13.40 & 14.43 & 13.46 & 14.01 & 13.80 & 12.84 & 13.77 \\
\textsc{it} & 24.42 & 23.89 & 23.96 & 24.34 & 24.76 & 22.86 & 23.96 & 24.54 \\
\textsc{lt} & 13.15 & 13.00 & 12.94 & 12.61 & 12.57 & 12.78 & 11.65 & 12.84 \\
\textsc{lv} & 14.43 & 13.84 & 14.75 & 14.17 & 14.17 & 13.65 & 13.16 & 14.19 \\
\textsc{nl} & 23.42 & 22.85 & 22.02 & 21.92 & 22.20 & 21.52 & 20.86 & 22.94 \\
\textsc{pl} & 20.01 & 19.27 & 19.19 & 19.36 & 19.95 & 19.42 & 18.94 & 20.09 \\
\textsc{pt} & 27.33 & 26.58 & 26.55 & 27.26 & 27.21 & 26.80 & 24.78 & 27.23 \\
\textsc{ro} & 27.81 & 27.77 & 27.73 & 27.51 & 27.72 & 27.43 & 26.93 & 28.06 \\
\textsc{sk} & 20.53 & 21.11 & 20.18 & 20.21 & 20.59 & 19.80 & 19.44 & 20.59 \\
\textsc{sl} & 18.47 & 18.76 & 18.20 & 17.67 & 18.50 & 17.57 & 17.55 & 18.53 \\
\textsc{sv} & 24.17 & 24.21 & 22.89 & 23.32 & 23.39 & 22.72 & 22.09 & 23.43 \\
\bottomrule
\end{tabular}
}

    \caption{sBLEU scores of Gemma-3-4B  per target language. Each language has its performance variance significantly explained by the party variable  according to a Kruskal-Wallis test ($p<0.01$).}
    \label{tab:sBLEU_Gemma-3-4B_per_tgt}
\end{table}

\begin{table}[h]
\resizebox{\columnwidth}{!}{
\begin{tabular}{crrrrrrrr}
\toprule
tgt & ALDE & ECR & EFA & EFD & EPP & NA & NGL & S\&D \\
\midrule
\textsc{bg} & 15.90 & 16.75 & 15.89 & 16.19 & 16.44 & 14.75 & 15.85 & 16.65 \\
\textsc{cs} & 12.60 & 12.46 & 12.44 & 12.39 & 12.69 & 12.04 & 11.56 & 12.76 \\
\textsc{da} & 15.96 & 15.78 & 15.50 & 15.09 & 15.74 & 14.37 & 15.09 & 15.61 \\
\textsc{de} & 17.59 & 17.77 & 16.39 & 17.36 & 17.59 & 14.67 & 16.26 & 17.60 \\
\textsc{el} & 12.51 & 12.31 & 12.25 & 12.34 & 12.93 & 11.41 & 11.57 & 12.92 \\
\textsc{en} & 33.19 & 32.89 & 31.86 & 33.34 & 34.01 & 31.88 & 32.21 & 33.93 \\
\textsc{es} & 26.73 & 25.83 & 25.66 & 27.33 & 28.22 & 25.99 & 26.49 & 27.46 \\
\textsc{et} & 5.90 & 5.81 & 6.09 & 5.42 & 5.56 & 5.63 & 5.28 & 5.63 \\
\textsc{fi} & 8.50 & 8.11 & 8.43 & 7.69 & 8.11 & 7.73 & 7.76 & 8.07 \\
\textsc{fr} & 22.26 & 21.90 & 21.82 & 22.77 & 23.61 & 21.05 & 22.40 & 23.05 \\
\textsc{hu} & 8.78 & 8.28 & 9.09 & 8.60 & 8.99 & 8.57 & 8.10 & 8.72 \\
\textsc{it} & 18.23 & 17.15 & 17.85 & 17.95 & 18.60 & 17.08 & 17.91 & 18.42 \\
\textsc{lt} & 8.35 & 8.01 & 8.44 & 7.99 & 8.08 & 7.96 & 7.62 & 8.05 \\
\textsc{lv} & 7.57 & 7.30 & 7.88 & 7.21 & 7.40 & 7.13 & 6.85 & 7.30 \\
\textsc{nl} & 15.98 & 15.61 & 15.03 & 15.36 & 15.51 & 14.78 & 14.65 & 15.76 \\
\textsc{pl} & 12.10 & 11.61 & 11.87 & 11.85 & 12.29 & 11.78 & 11.68 & 12.19 \\
\textsc{pt} & 20.93 & 20.02 & 20.40 & 21.01 & 21.19 & 20.34 & 19.09 & 21.12 \\
\textsc{ro} & 19.34 & 18.80 & 19.16 & 19.52 & 19.69 & 18.72 & 18.56 & 19.70 \\
\textsc{sk} & 9.29 & 9.18 & 9.24 & 8.97 & 9.31 & 8.74 & 8.68 & 9.38 \\
\textsc{sl} & 10.26 & 9.98 & 10.29 & 9.61 & 10.08 & 9.68 & 9.92 & 10.09 \\
\textsc{sv} & 15.22 & 14.70 & 14.29 & 14.52 & 14.78 & 13.59 & 14.20 & 14.84 \\
\bottomrule
\end{tabular}}
    \caption{sBLEU scores of Qwen3-4B  per target language. Each language has its performance variance significantly explained by the party variable  according to a Kruskal-Wallis test ($p<0.01$).}
    \label{tab:sBLEU_Qwen3-4B_per_tgt}
\end{table}

\begin{table}[h]
    \centering
    \resizebox{\columnwidth}{!}{
    \begin{tabular}{crrrrrrrr}
\toprule
tgt & ALDE & ECR & EFA & EFD & EPP & NA & NGL & S\&D \\
\midrule
\textsc{bg} & 20.83 & 21.56 & 21.28 & 21.59 & 21.81 & 19.80 & 20.88 & 21.93 \\
\textsc{cs} & 17.21 & 16.90 & 16.80 & 17.29 & 17.37 & 16.47 & 15.99 & 17.46 \\
\textsc{da} & 20.91 & 20.35 & 20.60 & 19.88 & 20.55 & 19.27 & 19.93 & 20.52 \\
\textsc{de} & 21.35 & 21.73 & 20.38 & 21.00 & 21.10 & 18.10 & 19.70 & 21.34 \\
\textsc{el} & 18.70 & 18.19 & 18.21 & 18.16 & 18.89 & 17.29 & 17.27 & 19.08 \\
\textsc{en} & 36.04 & 35.65 & 34.67 & 36.00 & 36.35 & 34.65 & 35.01 & 36.60 \\
\textsc{es} & 30.81 & 30.05 & 29.85 & 31.39 & 32.20 & 30.22 & 30.66 & 31.52 \\
\textsc{et} & 8.62 & 8.27 & 8.67 & 7.91 & 8.16 & 8.14 & 7.66 & 8.19 \\
\textsc{fi} & 12.01 & 11.42 & 11.73 & 10.92 & 11.45 & 10.77 & 10.80 & 11.47 \\
\textsc{fr} & 25.99 & 25.45 & 25.35 & 26.27 & 27.19 & 24.74 & 25.68 & 26.78 \\
\textsc{hu} & 12.39 & 11.73 & 12.82 & 11.97 & 12.37 & 12.19 & 11.55 & 12.19 \\
\textsc{it} & 21.27 & 20.15 & 20.70 & 21.23 & 21.78 & 19.94 & 20.77 & 21.50 \\
\textsc{lt} & 11.86 & 11.10 & 11.77 & 11.40 & 11.41 & 11.26 & 10.55 & 11.43 \\
\textsc{lv} & 11.91 & 11.09 & 11.88 & 11.63 & 11.87 & 11.24 & 10.75 & 11.59 \\
\textsc{nl} & 20.03 & 19.54 & 19.05 & 19.03 & 19.34 & 18.89 & 18.41 & 19.79 \\
\textsc{pl} & 16.12 & 15.39 & 15.79 & 15.59 & 16.20 & 15.76 & 15.22 & 16.21 \\
\textsc{pt} & 24.34 & 23.47 & 23.64 & 24.41 & 24.60 & 23.76 & 22.26 & 24.55 \\
\textsc{ro} & 23.91 & 23.07 & 23.58 & 23.40 & 23.86 & 23.09 & 22.81 & 24.13 \\
\textsc{sk} & 17.03 & 16.99 & 16.46 & 16.50 & 17.00 & 16.15 & 16.08 & 17.07 \\
\textsc{sl} & 14.57 & 14.30 & 14.28 & 13.93 & 14.64 & 13.88 & 13.96 & 14.50 \\
\textsc{sv} & 19.85 & 19.57 & 18.89 & 18.83 & 19.18 & 18.43 & 18.43 & 19.27 \\
\bottomrule
\end{tabular}}
    \caption{sBLEU scores of Qwen3-8B  per target language. Each target language has its performance variance significantly explained by the party variable  according to a Kruskal-Wallis test ($p<0.01$).}
    \label{tab:sBLEU_qwen3-8b_per_tgt}
\end{table}

\begin{table}[h]
\resizebox{\columnwidth}{!}{
    \begin{tabular}{crrrrrrrr}
\toprule
tgt & ALDE & ECR & EFA & EFD & EPP & NA & NGL & S\&D \\
\midrule
\textsc{bg} & 18.13 & 19.00 & 18.10 & 18.60 & 18.39 & 17.51 & 17.94 & 18.73 \\
\textsc{cs} & 17.75 & 17.78 & 17.58 & 17.70 & 17.90 & 17.29 & 16.89 & 18.00 \\
\textsc{da} & 19.80 & 19.61 & 19.54 & 18.45 & 19.17 & 18.50 & 18.53 & 19.22 \\
\textsc{de} & 20.86 & 21.23 & 20.12 & 20.33 & 20.34 & 18.11 & 19.22 & 20.55 \\
\textsc{el} & 17.90 & 17.73 & 17.11 & 17.54 & 18.10 & 16.55 & 16.49 & 18.20 \\
\textsc{en} & 35.52 & 35.23 & 34.52 & 35.94 & 35.80 & 34.27 & 34.39 & 36.06 \\
\textsc{es} & 29.52 & 28.84 & 28.80 & 30.26 & 30.87 & 29.15 & 29.48 & 30.21 \\
\textsc{et} & 8.16 & 8.02 & 8.30 & 7.53 & 7.75 & 7.99 & 7.39 & 7.86 \\
\textsc{fi} & 10.70 & 10.61 & 10.84 & 9.78 & 10.23 & 10.17 & 9.78 & 10.27 \\
\textsc{fr} & 25.23 & 24.64 & 24.86 & 25.58 & 26.33 & 24.04 & 25.27 & 25.89 \\
\textsc{hu} & 12.63 & 12.52 & 13.12 & 12.13 & 12.56 & 12.68 & 11.84 & 12.51 \\
\textsc{it} & 20.92 & 19.92 & 20.50 & 20.50 & 21.10 & 19.80 & 20.59 & 21.00 \\
\textsc{lt} & 8.91 & 8.57 & 9.05 & 8.53 & 8.64 & 8.71 & 8.01 & 8.73 \\
\textsc{lv} & 9.04 & 8.81 & 9.12 & 8.80 & 8.98 & 8.81 & 8.40 & 8.96 \\
\textsc{nl} & 20.50 & 20.16 & 19.72 & 19.38 & 19.63 & 19.19 & 18.54 & 20.02 \\
\textsc{pl} & 15.95 & 15.66 & 15.65 & 15.71 & 16.22 & 15.70 & 15.21 & 16.34 \\
\textsc{pt} & 23.83 & 23.16 & 23.27 & 23.79 & 23.97 & 23.48 & 21.95 & 23.84 \\
\textsc{ro} & 21.75 & 21.46 & 21.68 & 21.74 & 22.04 & 21.55 & 21.09 & 22.15 \\
\textsc{sk} & 13.70 & 13.62 & 13.42 & 13.36 & 13.65 & 13.21 & 13.00 & 13.84 \\
\textsc{sl} & 13.57 & 13.65 & 13.56 & 13.16 & 13.70 & 13.10 & 13.12 & 13.80 \\
\textsc{sv} & 21.08 & 20.89 & 20.02 & 20.09 & 20.40 & 20.00 & 19.50 & 20.56 \\
\bottomrule
\end{tabular}}
    \caption{sBLEU scores of Llama-3.1-8B  per target language. Each language has its performance variance significantly explained by the party variable  according to a Kruskal-Wallis test ($p<0.01$).}
    \label{tab:sBLEU_Llama-3.1-8B-Instruct_per_tgt}
\end{table}

}

{}

\todos

\end{document}